\documentclass[letterpaper]{article}
\usepackage[preprint]{aaai2027}
\usepackage[hyphens]{url}
\usepackage{graphicx}
\usepackage{natbib}
\usepackage{caption}
\usepackage{amsmath,mathtools}
\usepackage{amssymb}
\usepackage{bm}
\usepackage{booktabs}
\usepackage{makecell}
\usepackage{xspace}

\newcommand{\methodname}{WDR\xspace}

\newcommand{\wdrtablestyle}{%
  \small
}

\title{Towards Valid B-Rep Generation: Training-Free Wireframe Anomaly Detection and Repair}
\author{
    Jingyu Wu,
    Youcheng Cai\corresponding,
    Tengyu Luo,
    Ligang Liu
}
\affiliations{
    University of Science and Technology of China, Hefei, China\\
    andyng@mail.ustc.edu.cn, caiyoucheng@ustc.edu.cn,\\
    luoty060401@mail.ustc.edu.cn, lgliu@ustc.edu.cn
}

\begin{document}
\maketitle

\begin{abstract}
Multi-stage boundary representation (B-Rep) generation leverages intermediate wireframes to synthesize CAD models. However, geometric and topological risks in these wireframes---such as self-intersections, edge collapses, and disconnected vertices---can propagate to invalid final B-Reps. Mitigating such failures by retraining large generative models is computationally prohibitive. We propose Wireframe Detection and Repair (WDR), a training-free framework that intervenes at the intermediate wireframe stage to improve downstream B-Rep validity. WDR features a Geometric-Topology Anomaly Detector (GTAD) that combines parallel VLM-based coarse screening with geometric and topological detectors to predict downstream invalidity risk and route generation to dedicated branches. An Energy-Guided Geometric-Topology Repair (EGGTR) module then performs detector-triggered guided regeneration through geometry and topology branches. By scaling test-time computation via Energy-Guided Resampling and training-free guidance for diffusion models, WDR can be integrated into autoregressive and diffusion pipelines without retraining. Extensive experiments demonstrate consistent improvements in kernel-checked validity while largely retaining the measured diversity and distributional quality of synthesized CAD models. The code will be made publicly available upon acceptance.
\end{abstract}

\section{Introduction}
\label{sec:intro}
 
Boundary Representation (B-Rep) \cite{weiler1986topological} serves as the foundational representation in modern Computer-Aided Design (CAD) systems, preserving the structural properties required for downstream applications, including parametric editing, physical simulation, and precision manufacturing. Recently, learning-based B-Rep synthesis has emerged as a promising frontier to automate CAD modeling. While early command-based approaches \cite{wu2021deepcad, xu2022skexgen, xu2023hierarchical} address this problem by predicting sequences of modeling operations, their applicability is hindered by the relative scarcity of datasets containing complete construction histories.
 
Consequently, attention has been directed toward direct B-Rep generation methods. Single-stage approaches \cite{lee2025brepdiff, liu2025hola, xu2025autobrep} aim to learn B-Reps within a unified framework, imposing a substantial burden on the learning process that often results in training instability and limited flexibility. Alternatively, multi-stage approaches \cite{jayaraman2023solidgen, xu2024brepgen, li2025dtgbrepgen, li2025stitchashape, brepforge2026, qin2026brepgd} offer greater adaptability by decomposing the generation problem. These methods typically employ separate generators for different primitives and produce intermediate wireframes, which encode rich geometric and topological information, thereby handling complex structural constraints.
\begin{figure}[t]
  \centering
  \includegraphics[width=\linewidth]{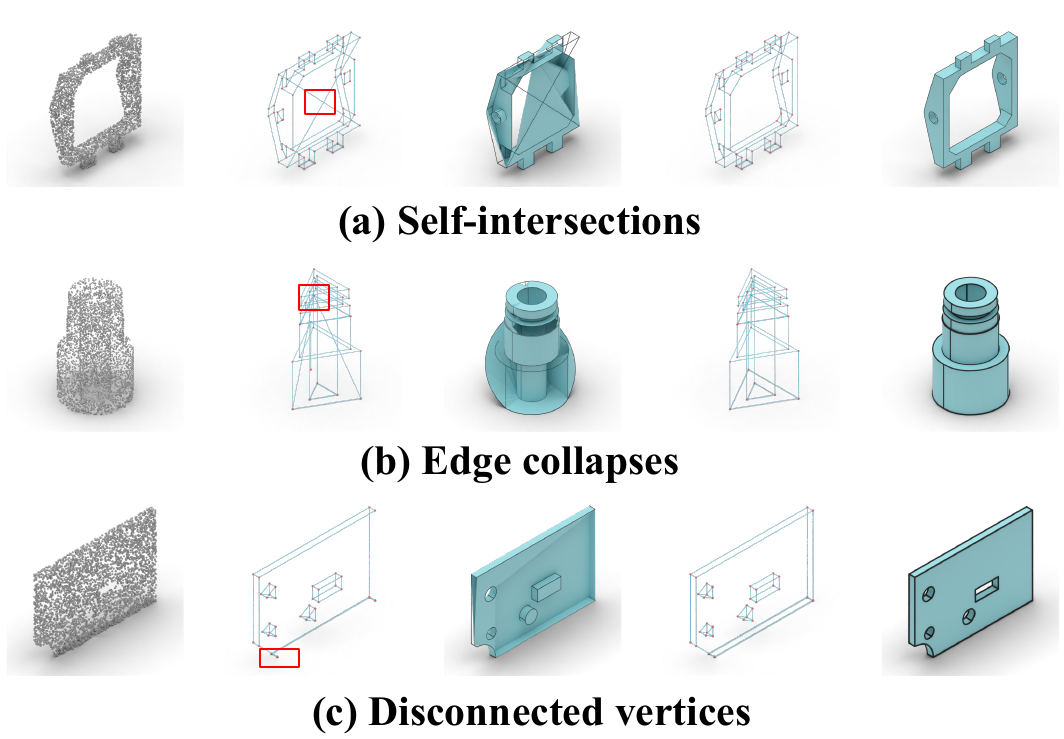}
\caption{Representative WDR-guided regeneration results. From left to right, each row shows the input point cloud, the initial wireframe and its generated B-Rep, and the WDR-guided wireframe and corresponding final B-Rep. Red boxes mark (a) self-intersections, (b) edge collapses, and (c) disconnected vertices.}
  \label{fig:intro}
\end{figure}

However, the majority of existing multi-stage approaches focus primarily on the visual quality of the final B-Reps, often overlooking structural risks in intermediate wireframes. Consequently, these pipelines are particularly vulnerable to error accumulation \cite{brepforge2026,xu2025autobrep,liu2025hola}. Such failures can stem from geometric and topological anomalies---including self-intersections, edge collapses, and disconnected vertices---arising in intermediate wireframes, as shown in Fig.~\ref{fig:intro}. A straightforward solution might involve fine-tuning existing generative models to mitigate these specific errors. However, retraining large models is both computationally intensive and data-intensive. This raises a critical question: \textit{Can a training-free plug-and-play module reduce downstream invalidity by intervening at the intermediate wireframe stage?} The primary challenges are (1) identifying complementary risk signals that predict downstream checker failure and (2) using these signals to guide regeneration without modifying the generator weights.

To address these challenges, we propose a training-free plug-and-play module, termed WDR (Wireframe Detection and Repair), that improves the validity of B-Rep generation by explicitly regularizing intermediate wireframe geometry and topology. Specifically, we introduce a \textbf{Geometric-Topology Anomaly Detector (GTAD)} that combines a VLM-based coarse screening signal with geometric and topological detectors to estimate complementary downstream-invalidity risks. Next, our \textbf{Energy-Guided Geometric-Topology Repair (EGGTR)} module performs branch-specific guided regeneration: its geometry branch biases regenerated coordinates away from high-risk configurations, while its topology branch reranks regenerated connectivity variables to reduce discrete constraint violations. We propose Energy-Guided Resampling for autoregressive generation and extend the framework to diffusion-based geometry generation through training-free guidance. Experiments with DTGBrepGen \cite{li2025dtgbrepgen}, Stitch-A-Shape \cite{li2025stitchashape}, and BrepForge \cite{brepforge2026} show that WDR improves kernel-checked validity without retraining the generators.
 
In summary, our core contributions are as follows:
\begin{itemize}
    \item We present a \textbf{training-free plug-and-play wireframe detection and repair framework} that operates on intermediate wireframes within multi-stage B-Rep generation pipelines and improves output validity without retraining the generators.
    
    \item We propose a \textbf{Geometric-Topology Anomaly Detector (GTAD)} that combines complementary VLM-based coarse screening with geometric and topological energy-based detection in intermediate wireframes.
    
    \item We develop an \textbf{Energy-Guided Geometric-Topology Repair (EGGTR)} module with geometry and topology branches that perform detector-triggered guided regeneration and local candidate reranking.
\end{itemize}

\section{Related Work}
\label{sec:related}

\subsection{CAD Command Generation}
CAD command generation focuses on learning-based methods that model shape creation as a sequential construction process composed of sketch-and-extrude operations. The generated command sequences can be executed by a solid modeling kernel to produce editable parametric CAD files. DeepCAD and subsequent sketch--extrude models \cite{wu2021deepcad, xu2022skexgen,xu2023hierarchical,guo2025cadtransn} learn such command sequences for editable CAD synthesis. However, these methods require construction history annotations and are constrained by the limited availability of datasets containing complete construction histories \cite{willis2020fusion, wu2021deepcad}.

\subsection{Direct B-Rep Generation}

Direct B-Rep methods synthesize continuous geometry together with discrete topology and can be broadly grouped into single-stage and multi-stage approaches.

\textbf{Single-stage methods.} These methods model geometry and topology holistically. HoLa~\cite{liu2025hola} learns a holistic primitive--topology latent space; AutoBrep and BrepGPT~\cite{xu2025autobrep,li2025brepgpt} serialize both structures; and BrepDiff~\cite{lee2025brepdiff} denoises masked face grids.

\textbf{Multi-stage methods.} ComplexGen~\cite{guo2022complexgen} first predicts B-Rep primitives and their incidences from point clouds, then recovers a valid chain complex through constrained global optimization. SolidGen~\cite{jayaraman2023solidgen} generates vertices, edges, and faces; BrepGen~\cite{xu2024brepgen} recovers topology from structured latent geometry; DTGBrepGen~\cite{li2025dtgbrepgen} decouples adjacency and geometry; and Stitch-A-Shape and BrepForge~\cite{li2025stitchashape,brepforge2026} progressively assemble B-Reps through intermediate wireframes. Such stage-wise dependencies can propagate wireframe errors, whereas \methodname uses exposed wireframes to trigger guided regeneration at inference time without retraining the generator.

\subsection{Training-Free Guidance for Diffusion and Autoregressive Models}

Classifier-free guidance \cite{ho2022classifier} guides diffusion sampling by combining conditional and unconditional scores, while Loss-Guided Diffusion (LGD) \cite{song2023loss} enables plug-and-play control through gradients from an external loss. TFG \cite{ye2024tfg} further unifies such training-free guidance in diffusion spaces, and GOOD \cite{gao2026good} extends guided sampling to robust out-of-distribution detection.

For autoregressive models, training-free guidance has shifted from conditional token control \cite{keskar2019ctrl} toward test-time compute scaling. Recent works show that scaling inference computation can rival or surpass parameter scaling \cite{snell2024scalingllmtesttimecompute}, often through step-wise verification \cite{wang2024math} and confidence-calibrated generation \cite{chen2026rethinking}. In vision, GridAR \cite{park2025gridar} progressively steers autoregressive image decoding toward higher-quality outputs without retraining. Inspired by this paradigm, we introduce energy-guided resampling for direct B-Rep generation, using additional test-time computation to rerank local wireframe continuations without fine-tuning.

\begin{figure*}[t]
  \centering
  \includegraphics[width=0.9\textwidth]{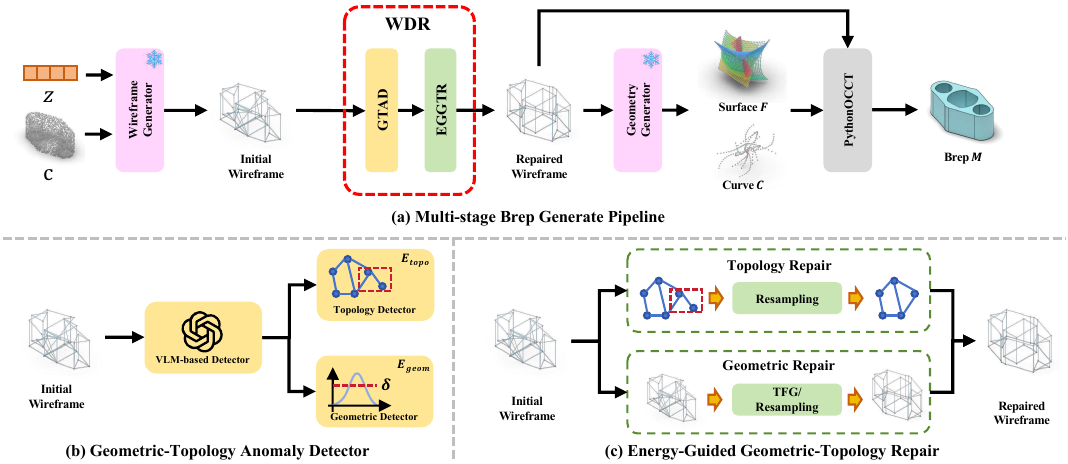}
\caption{
Overview of Wireframe Detection and Repair (WDR). (a) WDR plugs into pretrained multi-stage B-Rep generators at the wireframe generation stage, screens for downstream invalidity risk, and guides regeneration before geometry completion. (b) GTAD combines parallel VLM-based coarse screening with geometric and topological detectors. (c) EGGTR activates topology- and geometry-side guidance to obtain a revised wireframe for final B-Rep generation.
}
\label{fig:overview}
\end{figure*}
\section{Preliminaries}
\label{sec:preliminaries}
\subsection{Multi-stage B-Rep Generation}

Given latent noise $z$ and optional conditioning constraints $c$, multi-stage B-Rep generation approaches \cite{li2025dtgbrepgen, li2025stitchashape, brepforge2026} can be formulated as a sequential two-step process:
\begin{equation}
\mathcal{W} = g_{wire}(z, c), \quad \mathcal{G} = g_{geom}(\mathcal{W}, c),
\end{equation}
where $g_{wire}$ and $g_{geom}$ denote the wireframe generator and the geometry generator, respectively. $\mathcal{W}$ represents the intermediate wireframe, which encapsulates the discrete topological structure, while $\mathcal{G}$ denotes the continuous geometric attributes. The final B-Rep model is constructed by integrating the topological relationships and geometric entities $\mathcal{M} = (\mathcal{W}, \mathcal{G})$ using the CAD kernel of OpenCASCADE Technology (OCCT) \cite{paviot2022pythonocc}.


Formally, we represent the intermediate wireframe as a topology-aware spatial structure $\mathcal{W} = (\mathcal{V}, \mathcal{E}, \mathcal{T})$, where
\begin{equation}
\begin{aligned}
\mathcal{V} = \{v_i \in \mathbb{R}^3 \mid i=1,\cdots, n\}, \\ 
\mathcal{E} \subseteq \{(v_i, v_j) \in \mathcal{V} \times \mathcal{V} \mid i \neq j\}.
\end{aligned}
\end{equation}
Here, $\mathcal{V}$ contains 3D vertex coordinates, $\mathcal{E}$ contains their edge connectivity, and $\mathcal{T}$ stores the vertex--edge, edge--loop, loop--face, and face--shell relations together with ordered boundary information. This compact abstraction does not require different generators to share the same native tokenization; its complete definition and its prefix-wise realization are provided under \textit{Comparison of Sampling Strategies} in the Supplementary Material. The geometric attributes are $\mathcal{G} = (\mathcal{C}, \mathcal{F})$, where $\mathcal{C}$ represents the parametric curves corresponding to edges in $\mathcal{E}$, and $\mathcal{F}$ represents bounded faces defined by their boundary wires and underlying surfaces.

We focus on geometric and topological risks that arise during the wireframe generation stage ($g_{wire}$). WDR screens the available intermediate structure and triggers guided regeneration before these risks propagate to geometry generation ($g_{geom}$), with the goal of improving final kernel-checked validity.

\subsection{Validity}
\label{subsec:validity-definition}
Following prior B-Rep generation protocols, \emph{Valid} denotes whether the final constructed model passes an OCCT/PythonOCC kernel checker for a closed, manifold solid \cite{paviot2022pythonocc}. Producing such models is challenging because predicted geometric entities and topological elements must align sufficiently for kernel construction. The checker assesses the following structural conditions:
\begin{itemize}
\item \textbf{Watertightness:} The final model must form exactly one closed, manifold solid, where every edge is shared by two adjacent faces and no open boundaries exist.
\item \textbf{Loop Integrity:} The trimming loops of each face must have consistent orientation, and their parametric curves must be free of self-intersections.
\item \textbf{Shell Consistency:} The bounding shell must contain no bad shell edges or free/open edges.
\end{itemize}
A generated model that fails any of these checks is counted as invalid under our evaluation metric. Such kernel-checked validity is an important structural criterion, but it does not by itself establish manufacturability, functional correctness, or fitness for downstream engineering use. We use intermediate geometric and topological risk signals to guide regeneration and reduce their propagation to downstream construction.

\section{Method}
\label{sec:method}

\subsection{Problem Statement}
Given latent noise $z$ and optional conditioning constraints $c$, the generation process of pre-trained multi-stage B-Rep models \cite{li2025dtgbrepgen, li2025stitchashape, brepforge2026} can be decomposed into wireframe generation and geometry generation. To improve structural validity without retraining these large-scale models, we introduce \methodname, a training-free plug-and-play wireframe detection-and-repair framework. We reformulate the generation process as:
\begin{equation}
\widetilde{\mathcal W}=F_{WDR}(g_{wire}(z,c)),\;
\mathcal M=(\widetilde{\mathcal W},g_{geom}(\widetilde{\mathcal W},c)).
\end{equation}
where $F_{WDR}$ represents our training-free Wireframe Detection and Repair (WDR) framework and $\widetilde{\mathcal W}$ is its guided wireframe output. Rather than acting as a post-processing filter over a completed B-Rep, $F_{WDR}$ screens the intermediate wireframe and guides its regeneration before subsequent geometry generation.

\subsection{Overview}
We present WDR, a training-free framework designed to enhance the kernel-checked validity of generated B-Rep models. The overall pipeline is illustrated in Fig.~\ref{fig:overview}. It operates in two coupled phases. First, the \textbf{Geometric-Topology Anomaly Detector (GTAD)} combines three parallel signals to predict whether an intermediate wireframe is at risk of producing a checker-invalid final B-Rep and to select active guidance branches. Second, the \textbf{Energy-Guided Geometric-Topology Repair (EGGTR)} module guides regeneration using geometry- and topology-specific objectives without altering the generator weights. Here, ``repair'' refers to detector-triggered guided regeneration---local candidate reranking for autoregressive stages or training-free guidance for diffusion stages---rather than deterministic editing of a fixed, completed wireframe.

\subsection{Geometric-Topology Anomaly Detector}
\label{subsec:GTAD}
The Geometric-Topology Anomaly Detector (GTAD) estimates downstream invalidity risk from the intermediate wireframe $\mathcal{W}$. We design a parallel, multimodal mechanism comprising a Vision-Language Model (VLM)-based detector, a geometric detector, and a topological detector. For its reported evaluation, the positive class is an unguided sample whose final B-Rep fails the OCCT-based checker. Thus, GTAD F1 measures downstream invalidity-risk prediction rather than direct ground-truth classification of every wireframe anomaly.

\textbf{VLM-based Detector:} Recent advances have demonstrated the capabilities of vision-language models (VLMs) in visual understanding and spatial reasoning \cite{hong2023threedllm,liu2023llava}. We employ a VLM as a parallel coarse-screening signal complementary to the geometric and topological detectors. Specifically, $R$ takes multi-view renderings of $\mathcal{W}$ as input and returns a binary risk signal $R(\mathcal{W}) \in \{0,1\}$. It can flag visually apparent structural risks, such as severe crossings, collapses, inconsistent connectivity, or disconnected components, but neither certifies final CAD validity nor supplies a scalar repair energy. A positive VLM signal serves only as a routing trigger for both guidance branches. In our implementation, we utilize Qwen3.5-Flash~\cite{qwen35blog}. Backbone and latency analyses are provided under \textit{Analysis on GTAD VLM-based Detector}, and the complete prompt is provided under \textit{VLM Detector Prompt}, in the Supplementary Material.

\textbf{Geometric Detector:} To analyze self-intersection-related geometric anomalies in the wireframe, we introduce an energy-based geometric detector. Following prior work~\cite{yu2021repulsive}, we compute the geometry energy of the wireframe using discrete tangent-point energy:
\begin{equation}
E_{\mathrm{geom}}(\mathcal{W})=\frac{1}{|\mathcal{E}|}\sum_{e\in\mathcal{E}} E_{tpe}(e),
\end{equation}
where $E_{tpe}(e)$ denotes the discrete tangent-point energy for edge $e$, and $|\mathcal{E}|$ is the number of wireframe edges. In its continuous setting, tangent-point energy is a nonlocal self-avoidance energy whose behavior distinguishes an approaching self-intersection from points that are close both spatially and along a curve~\cite{yu2021repulsive}. Our detector uses a finite discrete approximation, so the aggregated value is treated only as a heuristic self-intersection-related risk score, not as a complete predicate for intersections, edge collapses, or legal near-contact. We set $\delta=7.0$ based on the threshold-sensitivity study reported under \textit{Analysis on GTAD Geometric Detector} in the Supplementary Material; if $E_{\mathrm{geom}}(\mathcal{W}) > \delta$, the geometry branch is activated. Controlled cases covering interior crossings, near-collapsed edges, and legal near-contact are reported in the same supplementary section.

\textbf{Topological Detector:} For closed, manifold B-Rep models, where each boundary edge is shared by exactly two distinct faces, we use four validity-inspired structural checks: the \textit{Vertex-Edge}, \textit{Edge-Loop}, \textit{Loop-Face}, and \textit{Face-Shell} constraints. These checks assess local incidence, loop integrity, trimming-loop consistency, and shell connectivity, respectively. In particular, the degree bound and minimum loop size are representation-calibrated risk heuristics for the evaluated serializers and data, rather than universal necessary conditions for every legal B-Rep. Let $r_i(\mathcal{W})$ denote the violation score of constraint $C_i$. We define the topological energy as
\begin{equation}
E_{\mathrm{topo}}(\mathcal{W})=\sum_{i=1}^4 r_i(\mathcal{W}).
\end{equation}
We activate the topology branch whenever $E_{\mathrm{topo}}(\mathcal{W})>0$. Detailed definitions and analyses of these constraints are provided under \textit{Analysis on GTAD Topology Detector} in the Supplementary Material.

In practice, GTAD runs the three detectors in parallel. Because the VLM does not attribute a specific anomaly type, a positive VLM signal activates both guidance branches. A positive geometric or topological signal activates its corresponding branch, and multiple positive signals jointly determine the active set. The wireframe is accepted without guided regeneration only when all three detectors return negative. The detailed routing workflow is provided under \textit{GTAD Routing} in the Supplementary Material.

\subsection{Energy-Guided Geometric-Topology Repair}
\label{subsec:EGGTR}
Once GTAD activates one or more branches, the Energy-Guided Geometric-Topology Repair (EGGTR) module performs targeted test-time guided regeneration before downstream B-Rep construction. The geometry branch biases regenerated continuous coordinates toward lower geometric risk, whereas the topology branch reranks regenerated connectivity variables according to loop- and shell-related constraints. Neither branch deterministically edits a fixed completed wireframe. Since autoregressive B-Rep generators serialize geometry and topology into structured token sequences for next-token prediction~\cite{jayaraman2023solidgen,li2025brepgpt,li2025stitchashape,xu2025autobrep,brepforge2026}, EGGTR scales test-time computation in this discrete setting; it also supports diffusion-based geometry generation.

\subsubsection{Energy-Guided Resampling via Local Candidate Reranking.}
For autoregressive generators, EGGTR performs detector-triggered regeneration under the same condition $c$. A straightforward baseline is Best-of-$N$ reranking, which evaluates structural energy only after generating $N$ complete sequences~\cite{collins_koo_2005_discriminative,snell2024scalingllmtesttimecompute}. Inspired by progressive test-time scaling~\cite{park2025gridar}, we instead apply local energy-aware candidate reranking during decoding, retaining the accepted prefix while rejecting unfavorable continuations before completing the wireframe.

Specifically, the pre-trained autoregressive model $f_{\phi}$ performs next-token prediction and factorizes the conditional sequence likelihood as
\begin{equation}
f_{\phi}(\mathbf{x}\mid c)
=
\prod_{t=1}^{T}
f_{\phi}(x_t\mid\mathbf{x}_{<t},c),
\end{equation}
where $\mathbf{x}=(x_1, \dots, x_T)$ is a token sequence of length $T$, $c$ is the condition, and $\phi$ denotes the model parameters. The mapping $\mathcal W(\cdot)$ converts a complete sequence to a wireframe, and $E$ denotes $E_{\mathrm{geom}}$, $E_{\mathrm{topo}}$, or their sum according to the active guidance branches.

\textbf{Global motivation.} The following energy-tilted distribution provides a conceptual objective for favoring high-likelihood sequences with low structural energy:
\begin{equation}
q^{\star}(\mathbf{x}\mid c)
\propto
f_{\phi}(\mathbf{x}\mid c)
\exp\!\left[-E\!\left(\mathcal W(\mathbf{x})\right)\right].
\end{equation}
Our implementation does not sample exactly from this global distribution. Instead, it approximates the objective through a finite set of local candidates at each token position.

\textbf{Local candidate reranking.} Let $\mathcal W_t(\mathbf{x}_{\leq t})$ denote the topology-aware partial wireframe recoverable from the current prefix. For a candidate continuation $x_t$, we evaluate the local energy increment
\begin{equation}
\Delta E_t(x_t)
=
E\!\left(\mathcal W_t(\mathbf{x}_{<t},x_t)\right)
-E\!\left(\mathcal W_{t-1}(\mathbf{x}_{<t})\right),
\end{equation}
using only the primitives and relations that are available at that prefix. The complete prefix mapping and masked-energy construction are provided under \textit{Comparison of Sampling Strategies} in the Supplementary Material.

At each token position where an active energy is evaluable, we draw a set $\mathcal{C}^t$ of at most $K$ distinct candidates from the base next-token proposal and deterministically select
\begin{equation}
x_t \leftarrow \arg\min_{x \in \mathcal{C}^t}
\left[-\log f_{\phi}(x\mid\mathbf{x}_{<t},c)+\Delta E_t(x)\right].
\end{equation}
We set $K=4$. Conditional on the sampled candidate set, this is a deterministic, approximate local selection rather than exact sampling from $q^{\star}$. The likelihood term retains the generator preference, while the energy increment discourages locally higher-risk continuations in the active geometry and topology branches.

\subsubsection{Generalization to Diffusion Models.}
Because representative multi-stage generators such as DTGBrepGen~\cite{li2025dtgbrepgen} use diffusion for geometry generation, we apply TFG~\cite{ye2024tfg} only to the geometry route; discrete topology guidance remains autoregressive. The original generator condition $c$ is preserved throughout the reverse trajectory. With cumulative noise retention $\bar{\alpha}_t$, the conditional noise predictor gives the Tweedie estimate~\cite{efron2011tweedies}
\begin{equation}
    \hat x_{0,t} = \frac{x_t-\sqrt{1-\bar{\alpha}_t}\,\epsilon_{\theta}(x_t,t,c)}{\sqrt{\bar{\alpha}_t}}.
\end{equation}

We adapt TFG to tilt the conditional base distribution toward lower geometry energy:
\begin{equation}
  q_{\lambda}(x_0\mid c)
  \propto p_{\theta}(x_0\mid c)
  \exp\!\left[-\lambda_{\mathrm{geom}}
  E_{\mathrm{geom}}\!\left(\mathcal{W}(x_0)\right)\right].
  \label{eq:target}
\end{equation}
At each denoising step, the corresponding guided reverse update is
\begin{equation}
  x_{t-1}
  =\operatorname{Sample}_{\theta}\!\left(x_t,\hat x_{0,t},c,t\right)
  +\frac{\Delta_t}{\sqrt{\alpha_t}}
  +\sqrt{\bar{\alpha}_{t-1}}\Delta_0.
  \label{eq:tfg-guided-update}
\end{equation}
Here, $\operatorname{Sample}_{\theta}(\cdot)$ is the conditional DDIM~\cite{song2021ddim} reverse step, while $\Delta_t$ and $\Delta_0$ are the noisy-state and clean-estimate guidance terms, respectively. Their exact gradients, smoothing, inner iterations, recurrence, forward re-noising kernel, and schedules are provided under \textit{Complete TFG Procedure} in the Supplementary Material.

\section{Experiments}
\subsection{Experimental Setup}

\begin{table}[t]
  \centering
  \wdrtablestyle
  \begin{tabular*}{\columnwidth}{@{\extracolsep{\fill}}lcc@{}}
    \toprule
    Method
    & \makecell[c]{Valid\\Novel\\Unique}
    & \makecell[c]{COV\\MMD\\JSD} \\
    \midrule

    \multicolumn{3}{c}{\textbf{DeepCAD}} \\
    \midrule
    DeepCAD & 65.8/97.9/91.0 & \textbf{79.17}/1.42/3.76 \\
    BrepGen & 62.2/99.7/99.6 & 56.19/1.23/2.20 \\
    DTGBrepGen & 79.5/98.2/98.0 & 75.04/\underline{1.05}/\underline{1.04} \\
    Stitch-A-Shape & 84.2/99.7/98.2 & 67.97/1.19/1.62 \\
    BrepForge & 85.3/\textbf{99.9}/\textbf{99.7} & \underline{78.21}/1.08/\textbf{1.02} \\
    \cmidrule(lr){1-3}
    \makecell[l]{DTGBrepGen\\+WDR}
      & 95.6/99.6/99.0
      & 75.02/1.06/1.13 \\
    \makecell[l]{Stitch-A-Shape\\+WDR}
      & \underline{97.0}/99.4/98.4
      & 72.53/\textbf{1.02}/1.15 \\
    \makecell[l]{BrepForge\\+WDR}
      & \textbf{97.4}/\underline{99.8}/\underline{99.7}
      & 74.79/1.08/1.09 \\

    \midrule
    \multicolumn{3}{c}{\textbf{ABC}} \\
    \midrule
    BrepGen & 43.8/99.8/\textbf{99.6} & 52.31/\textbf{1.13}/1.82 \\
    DTGBrepGen & 68.1/\textbf{99.9}/\underline{99.4} & 72.08/\underline{1.22}/\textbf{1.12} \\
    BrepForge & 75.4/99.5/99.1 & \textbf{74.90}/1.26/1.74 \\
    Stitch-A-Shape & 56.6/99.6/98.6 & 64.88/1.47/1.93 \\
    \cmidrule(lr){1-3}
    \makecell[l]{DTGBrepGen\\+WDR}
      & \underline{83.9}/\underline{99.9}/99.3
      & 69.80/1.25/\underline{1.32} \\
    \makecell[l]{Stitch-A-Shape\\+WDR}
      & 83.5/99.6/99.3
      & 65.71/1.37/1.88 \\
    \makecell[l]{BrepForge\\+WDR}
      & \textbf{86.3}/99.6/99.1
      & \underline{73.84}/1.26/1.76 \\
    \bottomrule
  \end{tabular*}
  \caption{Unconditional B-Rep generation on DeepCAD and ABC. Slash-separated entries follow the metric order shown in each header. MMD and JSD are multiplied by $10^2$.}
  \label{tab:overall}
\end{table}

\paragraph{Datasets.}
We use DeepCAD~\cite{wu2021deepcad} and ABC~\cite{Koch_2019_CVPR_ABC} for unconditional generation and Furniture~\cite{xu2024brepgen} for class-conditioned generation. Following prior protocols~\cite{xu2024brepgen,li2025dtgbrepgen,brepforge2026}, unconditional evaluation uses 3,000 generated samples and 1,000 references.

\paragraph{Evaluation Metrics.}
Following prior B-Rep generation protocols~\cite{jayaraman2023solidgen,xu2024brepgen,li2025dtgbrepgen}, we report both CAD-level and distribution-level metrics. The CAD-level metrics include \emph{Valid}, \emph{Novel}, and \emph{Unique}, measuring the ratios of outputs that pass kernel-based B-Rep validity checks, are unseen in the training set, and are non-duplicated, respectively. The distribution-level metrics include Coverage (COV), Minimum Matching Distance (MMD), and Jensen--Shannon Divergence (JSD), which measure reference-set coverage, nearest-neighbor shape distance, and point-distribution discrepancy based on sampled surface points. For point-cloud-conditioned reconstruction, we follow BrepForge~\cite{brepforge2026} and report \emph{Valid}, CD, EMD, and F-Score over the same 3,000 test conditions for the baseline and WDR, using identical condition IDs and base random seeds without excluding failed reconstructions. The complete evaluation protocol is provided in the Supplementary Material.

\paragraph{Implementation Details.}
We integrate \methodname as a plug-and-play inference-time module at the exposed wireframe-producing stages of DTGBrepGen, Stitch-A-Shape, and BrepForge~\cite{li2025dtgbrepgen,li2025stitchashape,brepforge2026}. Energy-Guided Resampling is used for autoregressive stages, whereas TFG is applied only to diffusion-based geometry generation. The original generator weights and conditioning interfaces remain unchanged. All experiments are conducted on an NVIDIA A100 GPU. Generator-specific intervention points, guidance configurations, and evaluation protocols are provided under \textit{Implementation Details} in the Supplementary Material.

\begin{figure*}[t]
  \centering
  \includegraphics[width=0.9\textwidth]{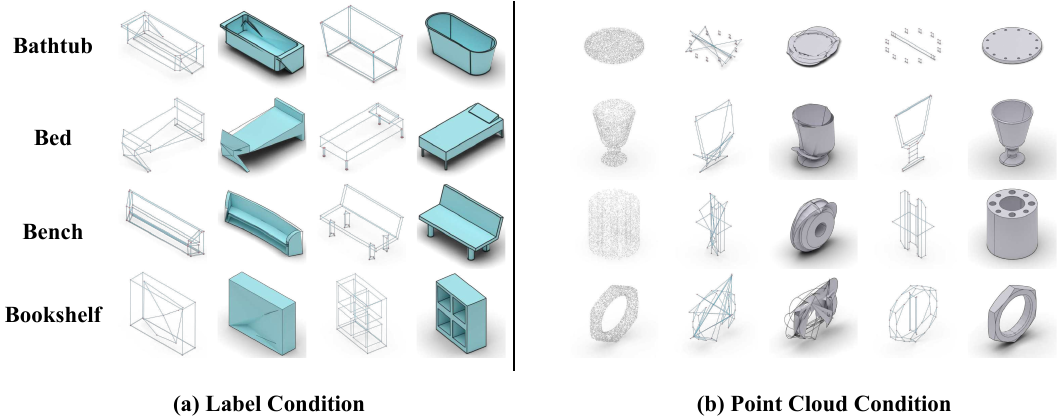}
  \caption{Qualitative results of conditioned B-Rep generation. Left: label-conditioned generation. Right: point-cloud-conditioned generation. In each panel, columns from left to right show the input condition (a class label or point cloud), the initial wireframe and its generated B-Rep, and the WDR-guided wireframe and corresponding B-Rep.}
  \label{fig:conditioned-generation}
\end{figure*}

\subsection{Quantitative Evaluation}

\paragraph{Unconditional Generation.}

Across DTGBrepGen, Stitch-A-Shape, and BrepForge on DeepCAD and ABC, \methodname improves \emph{Valid} by 10.9--26.9 percentage points while largely retaining \emph{Novel} and \emph{Unique} and comparable COV, MMD, and JSD (Tab.~\ref{tab:overall}). The consistent gains and limited distributional changes suggest that \methodname corrects risky intermediate structures without substantially narrowing the output support. Additional galleries are provided in the Supplementary Material.

\paragraph{Class-Conditioned Generation.}
On Furniture~\cite{xu2024brepgen}, \methodname improves the ten-class macro-average \emph{Valid} from 64.36\% to 73.63\% for DTGBrepGen and from 58.59\% to 68.04\% for Stitch-A-Shape (Tab.~\ref{tab:class-conditioned-average}). The similar absolute gains show that the intervention remains effective under semantic conditioning, while macro-averaging prevents frequent classes from dominating the comparison. Fig.~\ref{fig:conditioned-generation}(a) and the Supplementary Material provide representative and class-wise results.

\begin{table}[t]
  \centering
  \small
  \begin{tabular*}{\columnwidth}{@{}l@{\extracolsep{\fill}}cc@{}}
    \toprule
    Generator & w/o WDR & w/ WDR \\
    \midrule
    DTGBrepGen & 64.36 & \textbf{73.63} \\
    Stitch-A-Shape & 58.59 & \textbf{68.04} \\
    \bottomrule
  \end{tabular*}
  \caption{Ten-class macro-average \emph{Valid} scores (\%) for class-conditioned B-Rep generation on Furniture.}
  \label{tab:class-conditioned-average}
\end{table}

\begin{table}[t]
  \centering
  \small
  \begin{tabular*}{\columnwidth}{@{}l@{\extracolsep{\fill}}cccc@{}}
    \toprule
    Method & Valid(\%)$\uparrow$ & CD\(\downarrow\) & EMD\(\downarrow\)& F-Score\(\uparrow\) \\
    \midrule
    BrepForge & 87.1 & 0.93 & 2.45 & 0.91 \\
    BrepForge+WDR & 89.3(+2.2) & \textbf{0.87} & \textbf{2.19} & \textbf{0.94} \\
    \bottomrule
  \end{tabular*}
  \caption{Quantitative evaluation of point cloud-conditioned B-Rep reconstruction following BrepForge~\cite{brepforge2026}. Baseline and WDR use the same 3,000 conditions and base random seeds, and all metrics include failed reconstructions. Both CD and EMD scores are multiplied by $10^2$.}
  \label{tab:pc-conditioned}
\end{table}

\paragraph{Point-Cloud-Conditioned Generation.}
On 3,000 matched BrepForge conditions~\cite{brepforge2026}, \methodname improves \emph{Valid} from 87.1\% to 89.3\% together with CD, EMD, and F-Score (Tab.~\ref{tab:pc-conditioned}). In particular, CD and EMD decrease from 0.93 to 0.87 and from 2.45 to 2.19, while F-Score increases from 0.91 to 0.94. Thus, the additional kernel-valid outputs do not arise from trading away agreement with the input point cloud. Both variants share condition IDs and base seeds, and failed reconstructions remain in all metrics. Fig.~\ref{fig:conditioned-generation}(b) and the Supplementary Material provide qualitative results and protocol details.

\begin{figure}
  \centering
  \includegraphics[width=\linewidth]{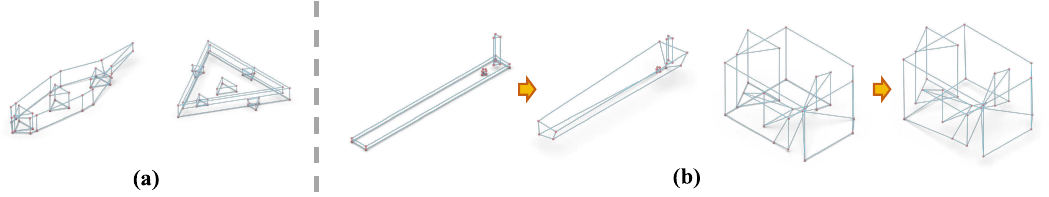}
  \caption{Failure cases. (a) GTAD misclassification. (b) EGGTR guided-regeneration failures.}
  \label{fig:failure-cases}
\end{figure}

\subsection{Ablation Study}
\label{subsec:ablation}
All ablations use DTGBrepGen~\cite{li2025dtgbrepgen} on ABC~\cite{Koch_2019_CVPR_ABC}.

\paragraph{Ablations on GTAD}
Combining all three detectors gives the best held-out downstream-risk F1 (Tab.~\ref{tab:gtad}). Its 81.97\% F1 exceeds every single detector and all two-signal variants, showing that visual, geometric, and topological evidence captures complementary failure patterns. Full routing reaches 83.9\% \emph{Valid}, only 0.4 points below always-on EGGTR, while reducing Valid$\rightarrow$Invalid regressions from 211 to 79; matched random routing reaches 80.2\%. These controls support selective intervention and are detailed in the Supplementary Material.

\begin{table}[t]
  \centering
  \small
  \begin{tabular*}{\columnwidth}{@{}l@{\extracolsep{\fill}}c@{}}
    \toprule
    Method & F1(\%)$\uparrow$ \\
    \midrule
    VLM  & 71.30 \\
    Geom. & 65.40 \\
    Topo. & 74.75 \\
    VLM + Geom. & 69.57 \\
    VLM + Topo. & 74.86 \\
    Geom. + Topo. & 69.65 \\
    VLM + Geom. + Topo. & \textbf{81.97} \\
    \bottomrule
  \end{tabular*}
  \caption{GTAD ablations on the held-out detector test cohort using F1-score for downstream invalidity-risk prediction. The positive class comprises unguided samples whose final B-Rep fails the OCCT-based checker.}
  \label{tab:gtad}
\end{table}

\paragraph{Ablations on EGGTR}
Geometry-only and topology-only regeneration improve \emph{Valid} by 4.8 and 9.3 points, respectively (Tab.~\ref{tab:ablation-main}). The larger topology-only gain identifies connectivity errors as the stronger bottleneck in this setting, whereas the best result from their combination shows that the two correction branches remain complementary. With $K=4$ local reranking, the combined variant reaches 83.9\% \emph{Valid}, 3.2 points above parallel Best-of-$4$~\cite{collins_koo_2005_discriminative,snell2024scalingllmtesttimecompute}, while retaining 99.9\% \emph{Novel} and 99.3\% \emph{Unique}. Selection and compute-budget details are provided in the Supplementary Material.

\begin{table}[t]
  \centering
  \small
  \begin{tabular*}{\columnwidth}{@{}l@{\extracolsep{\fill}}ccc@{}}
    \toprule
    Method & \makecell[c]{Valid\\(\%)$\uparrow$} & \makecell[c]{Novel\\(\%)$\uparrow$} & \makecell[c]{Unique\\(\%)$\uparrow$} \\
    \midrule
    baseline & 68.1 & \textbf{99.9} & \textbf{99.4} \\
    \midrule
    G.R. & \makecell{72.9\\(+4.8)} & 99.4 & 99.3 \\
    \midrule
    T.R. & \makecell{77.4\\(+9.3)} & 99.0 & 98.3 \\
    \midrule
    \makecell[l]{G.R. + T.R.\\(Rejection Sampling)} & \makecell{73.6\\(+5.5)} & 99.1 & 97.9 \\
    \midrule
    \makecell[l]{G.R. + T.R.\\(Best-of-$N$)} & \makecell{80.7\\(+12.6)} & 99.2 & 98.6 \\
    \midrule
    \makecell[l]{G.R. + T.R.\\(Our Resampling)} & \makecell{\textbf{83.9}\\(+15.8)} & 99.9 & 99.3 \\
    \bottomrule
  \end{tabular*}
  \caption{DTGBrepGen/ABC ablations. Parentheses give \emph{Valid} gains; Rejection Sampling allows one retry.}
  \label{tab:ablation-main}
\end{table}

\section{Conclusion}
We presented \methodname, a training-free framework that intervenes at exposed wireframes without changing generator weights. GTAD combines VLM, geometric, and topological risk signals, while EGGTR applies energy-guided autoregressive resampling or diffusion guidance. This separation couples diagnosis with localized correction: GTAD avoids regenerating every sample, and EGGTR focuses additional inference on the intermediate structures most likely to fail downstream construction.

Across three generators and two datasets, \methodname improves unconditional kernel-checked \emph{Valid} by 10.9--26.9 points while largely retaining measured diversity and distributional quality. It also improves class-conditioned validity for two generators and both validity and fidelity in point-cloud reconstruction. Together with the routing and regeneration ablations, these results show that complementary risk signals and targeted correction are both necessary.

\methodname optimizes an operational OCCT-based criterion rather than guaranteeing manufacturability or downstream functional performance. It adds inference cost, requires an exposed wireframe, and may encounter detector errors, unrecoverable corruptions (Fig.~\ref{fig:failure-cases}), or condition drift. Future work will explore efficient CAD-specific detectors, condition-aware guidance, adaptive stopping, and extensions to representations without explicit wireframes.
\bibliography{references}

\begin{thebibliography}{35}
\providecommand{\natexlab}[1]{#1}

\bibitem[{Chen et~al.(2025)Chen, Ravent\'{o}s, Cheng, Ganguli, and
  Druckmann}]{chen2026rethinking}
Chen, F.; Ravent\'{o}s, A.; Cheng, N.; Ganguli, S.; and Druckmann, S. 2025.
\newblock Rethinking Fine-Tuning when Scaling Test-Time Compute: Limiting
  Confidence Improves Mathematical Reasoning.
\newblock In \emph{Advances in Neural Information Processing Systems}.

\bibitem[{Collins and Koo(2005)}]{collins_koo_2005_discriminative}
Collins, M.; and Koo, T. 2005.
\newblock Discriminative Reranking for Natural Language Parsing.
\newblock \emph{Computational Linguistics}, 31(1): 25--70.

\bibitem[{Efron(2011)}]{efron2011tweedies}
Efron, B. 2011.
\newblock Tweedie's Formula and Selection Bias.
\newblock \emph{Journal of the American Statistical Association}, 106(496):
  1602--1614.

\bibitem[{Gao et~al.(2025)Gao, Liu, Li, Lyu, Gao, Yu, Xu, Wang, Shan, Liu, and
  Si}]{gao2026good}
Gao, X.; Liu, J.; Li, G.; Lyu, Y.; Gao, J.; Yu, W.; Xu, N.; Wang, L.; Shan, C.;
  Liu, Z.; and Si, C. 2025.
\newblock {GOOD}: Training-Free Guided Diffusion Sampling for
  Out-of-Distribution Detection.
\newblock In \emph{Advances in Neural Information Processing Systems}.

\bibitem[{Guo et~al.(2022)Guo, Liu, Pan, Liu, Tong, and
  Guo}]{guo2022complexgen}
Guo, H.; Liu, S.; Pan, H.; Liu, Y.; Tong, X.; and Guo, B. 2022.
\newblock {ComplexGen}: {CAD} Reconstruction by {B-Rep} Chain Complex
  Generation.
\newblock \emph{ACM Transactions on Graphics}, 41(4).

\bibitem[{Guo et~al.(2025)Guo, Dong, Cao, and Chen}]{guo2025cadtransn}
Guo, X.; Dong, X.; Cao, J.; and Chen, Z. 2025.
\newblock {CADTrans}: A Code Tree-Guided {CAD} Generative Transformer Model
  with Regularized Discrete Codebooks.
\newblock \emph{Graphical Models}, 139: 101262.

\bibitem[{Ho and Salimans(2022)}]{ho2022classifier}
Ho, J.; and Salimans, T. 2022.
\newblock Classifier-Free Diffusion Guidance.
\newblock arXiv:2207.12598.

\bibitem[{Hong et~al.(2023)Hong, Zhen, Chen, Zheng, Du, Chen, and
  Gan}]{hong2023threedllm}
Hong, Y.; Zhen, H.; Chen, P.; Zheng, S.; Du, Y.; Chen, Z.; and Gan, C. 2023.
\newblock {3D-LLM}: Injecting the {3D} World into Large Language Models.
\newblock In \emph{Advances in Neural Information Processing Systems},
  volume~36, 20482--20494.

\bibitem[{Jayaraman et~al.(2023)Jayaraman, Lambourne, Desai, Willis, Sanghi,
  and Morris}]{jayaraman2023solidgen}
Jayaraman, P.~K.; Lambourne, J.~G.; Desai, N.; Willis, K. D.~D.; Sanghi, A.;
  and Morris, N. J.~W. 2023.
\newblock SolidGen: An Autoregressive Model for Direct B-rep Synthesis.
\newblock arXiv:2203.13944.

\bibitem[{Keskar et~al.(2019)Keskar, McCann, Varshney, Xiong, and
  Socher}]{keskar2019ctrl}
Keskar, N.~S.; McCann, B.; Varshney, L.~R.; Xiong, C.; and Socher, R. 2019.
\newblock {CTRL}: A Conditional Transformer Language Model for Controllable
  Generation.
\newblock arXiv:1909.05858.

\bibitem[{Koch et~al.(2019)Koch, Matveev, Jiang, Williams, Artemov, Burnaev,
  Alexa, Zorin, and Panozzo}]{Koch_2019_CVPR_ABC}
Koch, S.; Matveev, A.; Jiang, Z.; Williams, F.; Artemov, A.; Burnaev, E.;
  Alexa, M.; Zorin, D.; and Panozzo, D. 2019.
\newblock {ABC}: A Big {CAD} Model Dataset for Geometric Deep Learning.
\newblock In \emph{Proceedings of the IEEE/CVF Conference on Computer Vision
  and Pattern Recognition}, 9601--9611.

\bibitem[{Lee et~al.(2025)Lee, Zhang, Jambon, and Kim}]{lee2025brepdiff}
Lee, M.; Zhang, D.; Jambon, C.; and Kim, Y.~M. 2025.
\newblock {BrepDiff}: Single-Stage {B}-rep Diffusion Model.
\newblock In \emph{Proceedings of the SIGGRAPH 2025 Conference Papers},
  SIGGRAPH Conference Papers '25.

\bibitem[{Li, Fu, and Chen(2025)}]{li2025dtgbrepgen}
Li, J.; Fu, Y.; and Chen, F. 2025.
\newblock {DTGBrepGen}: A Novel {B}-Rep Generative Model through Decoupling
  Topology and Geometry.
\newblock In \emph{Proceedings of the IEEE/CVF Conference on Computer Vision
  and Pattern Recognition}, 21438--21447.

\bibitem[{Li, Fu, and Chen(2026)}]{brepforge2026}
Li, J.; Fu, Y.; and Chen, F. 2026.
\newblock {BrepForge}: Factorized {B}-rep Synthesis via Wireframe Composition
  and Boundary-Conditioned Surface Instantiation.
\newblock In \emph{Proceedings of the SIGGRAPH 2026 Conference Papers}.

\bibitem[{Li et~al.(2025{\natexlab{a}})Li, Zhang, Chen, and
  Yan}]{li2025stitchashape}
Li, P.; Zhang, W.; Chen, J.; and Yan, D. 2025{\natexlab{a}}.
\newblock {Stitch-A-Shape}: Bottom-up Learning for {B}-Rep Generation.
\newblock In \emph{Proceedings of the SIGGRAPH 2025 Conference Papers},
  SIGGRAPH Conference Papers '25.

\bibitem[{Li et~al.(2025{\natexlab{b}})Li, Zhang, Quan, Zhang, Wonka, and
  Yan}]{li2025brepgpt}
Li, P.; Zhang, W.; Quan, W.; Zhang, B.; Wonka, P.; and Yan, D.-M.
  2025{\natexlab{b}}.
\newblock {BrepGPT}: Autoregressive {B}-rep Generation with Voronoi Half-Patch.
\newblock \emph{ACM Trans. Graph.}, 44(6): 226:1--226:18.

\bibitem[{Liu et~al.(2023)Liu, Li, Wu, and Lee}]{liu2023llava}
Liu, H.; Li, C.; Wu, Q.; and Lee, Y.~J. 2023.
\newblock Visual Instruction Tuning.
\newblock In \emph{Advances in Neural Information Processing Systems},
  volume~36, 34892--34916.

\bibitem[{Liu et~al.(2025)Liu, Xu, Yu, Xu, Cohen-Or, Zhang, and
  Huang}]{liu2025hola}
Liu, Y.; Xu, D.; Yu, X.; Xu, X.; Cohen-Or, D.; Zhang, H.; and Huang, H. 2025.
\newblock {HoLa}: {B}-Rep Generation using a Holistic Latent Representation.
\newblock \emph{ACM Trans. Graph.}, 44(4).

\bibitem[{Park et~al.(2026)Park, Jang, Kim, and Yang}]{park2025gridar}
Park, J.; Jang, H.; Kim, J.; and Yang, E. 2026.
\newblock Progress by Pieces: Test-Time Scaling for Autoregressive Image
  Generation.
\newblock In \emph{Proceedings of the IEEE/CVF Conference on Computer Vision
  and Pattern Recognition}.

\bibitem[{Paviot(2022)}]{paviot2022pythonocc}
Paviot, T. 2022.
\newblock {pythonocc}.
\newblock Python package for 3D CAD/BIM/PLM/CAM.

\bibitem[{Qin et~al.(2026)Qin, Luo, Hou, Fang, and Liu}]{qin2026brepgd}
Qin, F.; Luo, C.; Hou, J.; Fang, M.; and Liu, L. 2026.
\newblock {BRep-GD}: A Graph Diffusion Model for {CAD} Boundary Representation
  Generation.
\newblock \emph{IEEE Transactions on Visualization and Computer Graphics}, 32:
  1--17.

\bibitem[{{Qwen Team}(2026)}]{qwen35blog}
{Qwen Team}. 2026.
\newblock Qwen3.5: Accelerating Productivity with Native Multimodal Agents.

\bibitem[{Snell et~al.(2025)Snell, Lee, Xu, and
  Kumar}]{snell2024scalingllmtesttimecompute}
Snell, C.; Lee, J.; Xu, K.; and Kumar, A. 2025.
\newblock Scaling {LLM} Test-Time Compute Optimally can be More Effective than
  Scaling Model Parameters.
\newblock In \emph{International Conference on Learning Representations}.

\bibitem[{Song, Meng, and Ermon(2021)}]{song2021ddim}
Song, J.; Meng, C.; and Ermon, S. 2021.
\newblock Denoising Diffusion Implicit Models.
\newblock In \emph{International Conference on Learning Representations}.

\bibitem[{Song et~al.(2023)Song, Zhang, Yin, Mardani, Liu, Kautz, Chen, and
  Vahdat}]{song2023loss}
Song, J.; Zhang, Q.; Yin, H.; Mardani, M.; Liu, M.-Y.; Kautz, J.; Chen, Y.; and
  Vahdat, A. 2023.
\newblock Loss-Guided Diffusion Models for Plug-and-Play Controllable
  Generation.
\newblock In \emph{Proceedings of the International Conference on Machine
  Learning}, volume 202 of \emph{Proceedings of Machine Learning Research},
  32483--32498.

\bibitem[{Wang et~al.(2024)Wang, Li, Shao, Xu, Dai, Li, Chen, Wu, and
  Sui}]{wang2024math}
Wang, P.; Li, L.; Shao, Z.; Xu, R.; Dai, D.; Li, Y.; Chen, D.; Wu, Y.; and Sui,
  Z. 2024.
\newblock {Math-Shepherd}: Verify and Reinforce {LLM}s Step-by-Step without
  Human Annotations.
\newblock In \emph{Proceedings of the 62nd Annual Meeting of the Association
  for Computational Linguistics (Volume 1: Long Papers)}, 9426--9439.

\bibitem[{Weiler(1986)}]{weiler1986topological}
Weiler, K.~J. 1986.
\newblock \emph{Topological Structures for Geometric Modeling}.
\newblock Ph.D. thesis, Rensselaer Polytechnic Institute, Troy, NY.

\bibitem[{Willis et~al.(2021)Willis, Pu, Luo, Chu, Du, Lambourne, Solar-Lezama,
  and Matusik}]{willis2020fusion}
Willis, K. D.~D.; Pu, Y.; Luo, J.; Chu, H.; Du, T.; Lambourne, J.~G.;
  Solar-Lezama, A.; and Matusik, W. 2021.
\newblock Fusion 360 Gallery: A Dataset and Environment for Programmatic {CAD}
  Construction from Human Design Sequences.
\newblock \emph{ACM Trans. Graph.}, 40(4).

\bibitem[{Wu, Xiao, and Zheng(2021)}]{wu2021deepcad}
Wu, R.; Xiao, C.; and Zheng, C. 2021.
\newblock {DeepCAD}: A Deep Generative Network for Computer-Aided Design
  Models.
\newblock In \emph{Proceedings of the IEEE/CVF International Conference on
  Computer Vision}, 6772--6782.

\bibitem[{Xu et~al.(2025)Xu, Jayaraman, Lambourne, Liu, Malpure, and
  Meltzer}]{xu2025autobrep}
Xu, X.; Jayaraman, P.; Lambourne, J.; Liu, Y.; Malpure, D.; and Meltzer, P.
  2025.
\newblock {AutoBrep}: Autoregressive {B}-Rep Generation with Unified Topology
  and Geometry.
\newblock In \emph{Proceedings of the SIGGRAPH Asia 2025 Conference Papers}, SA
  Conference Papers '25.

\bibitem[{Xu et~al.(2023)Xu, Jayaraman, Lambourne, Willis, and
  Furukawa}]{xu2023hierarchical}
Xu, X.; Jayaraman, P.~K.; Lambourne, J.~G.; Willis, K. D.~D.; and Furukawa, Y.
  2023.
\newblock Hierarchical Neural Coding for Controllable {CAD} Model Generation.
\newblock In \emph{Proceedings of the International Conference on Machine
  Learning}, volume 202 of \emph{Proceedings of Machine Learning Research},
  38443--38461.

\bibitem[{Xu et~al.(2024)Xu, Lambourne, Jayaraman, Wang, Willis, and
  Furukawa}]{xu2024brepgen}
Xu, X.; Lambourne, J.; Jayaraman, P.; Wang, Z.; Willis, K.; and Furukawa, Y.
  2024.
\newblock {BrepGen}: A {B}-rep Generative Diffusion Model with Structured
  Latent Geometry.
\newblock \emph{ACM Trans. Graph.}, 43(4).

\bibitem[{Xu et~al.(2022)Xu, Willis, Lambourne, Cheng, Jayaraman, and
  Furukawa}]{xu2022skexgen}
Xu, X.; Willis, K. D.~D.; Lambourne, J.~G.; Cheng, C.-Y.; Jayaraman, P.~K.; and
  Furukawa, Y. 2022.
\newblock {SkexGen}: Autoregressive Generation of {CAD} Construction Sequences
  with Disentangled Codebooks.
\newblock In \emph{Proceedings of the International Conference on Machine
  Learning}, volume 162 of \emph{Proceedings of Machine Learning Research},
  24698--24724.

\bibitem[{Ye et~al.(2024)Ye, Lin, Han, Xu, Liu, Liang, Ma, Zou, and
  Ermon}]{ye2024tfg}
Ye, H.; Lin, H.; Han, J.; Xu, M.; Liu, S.; Liang, Y.; Ma, J.; Zou, J.~Y.; and
  Ermon, S. 2024.
\newblock {TFG}: Unified Training-Free Guidance for Diffusion Models.
\newblock In \emph{Advances in Neural Information Processing Systems},
  volume~37, 22370--22417.

\bibitem[{Yu, Schumacher, and Crane(2021)}]{yu2021repulsive}
Yu, C.; Schumacher, H.; and Crane, K. 2021.
\newblock Repulsive Curves.
\newblock \emph{ACM Trans. Graph.}, 40(2).

\end{thebibliography}


\begin{thebibliography}{22}
\providecommand{\natexlab}[1]{#1}

\bibitem[{Ansaldi, De~Floriani, and Falcidieno(1985)}]{ansaldi1985geometric}
Ansaldi, S.; De~Floriani, L.; and Falcidieno, B. 1985.
\newblock Geometric Modeling of Solid Objects by Using a Face Adjacency Graph
  Representation.
\newblock \emph{ACM SIGGRAPH Computer Graphics}, 19(3): 131--139.

\bibitem[{{Anthropic}(2025)}]{anthropic2025claudehaiku45}
{Anthropic}. 2025.
\newblock Introducing Claude Haiku 4.5.
\newblock Anthropic News.
\newblock Accessed: 2026-05-05.

\bibitem[{Collins and Koo(2005)}]{collins_koo_2005_discriminative}
Collins, M.; and Koo, T. 2005.
\newblock Discriminative Reranking for Natural Language Parsing.
\newblock \emph{Computational Linguistics}, 31(1): 25--70.

\bibitem[{Doshi and {Gemini Team}(2025)}]{google2026gemini3flash}
Doshi, T.; and {Gemini Team}. 2025.
\newblock Gemini 3 Flash: Frontier Intelligence Built for Speed.
\newblock Google Blog.
\newblock Accessed: 2026-05-05.

\bibitem[{Jayaraman et~al.(2023)Jayaraman, Lambourne, Desai, Willis, Sanghi,
  and Morris}]{jayaraman2023solidgen}
Jayaraman, P.~K.; Lambourne, J.~G.; Desai, N.; Willis, K. D.~D.; Sanghi, A.;
  and Morris, N. J.~W. 2023.
\newblock SolidGen: An Autoregressive Model for Direct B-rep Synthesis.
\newblock arXiv:2203.13944.

\bibitem[{Li, Fu, and Chen(2025)}]{li2025dtgbrepgen}
Li, J.; Fu, Y.; and Chen, F. 2025.
\newblock {DTGBrepGen}: A Novel {B}-Rep Generative Model through Decoupling
  Topology and Geometry.
\newblock In \emph{Proceedings of the IEEE/CVF Conference on Computer Vision
  and Pattern Recognition}, 21438--21447.

\bibitem[{Li, Fu, and Chen(2026)}]{brepforge2026}
Li, J.; Fu, Y.; and Chen, F. 2026.
\newblock {BrepForge}: Factorized {B}-rep Synthesis via Wireframe Composition
  and Boundary-Conditioned Surface Instantiation.
\newblock In \emph{Proceedings of the SIGGRAPH 2026 Conference Papers}.

\bibitem[{Li et~al.(2025)Li, Zhang, Chen, and Yan}]{li2025stitchashape}
Li, P.; Zhang, W.; Chen, J.; and Yan, D. 2025.
\newblock {Stitch-A-Shape}: Bottom-up Learning for {B}-Rep Generation.
\newblock In \emph{Proceedings of the SIGGRAPH 2025 Conference Papers},
  SIGGRAPH Conference Papers '25.

\bibitem[{MacKay(2003)}]{mackay2003information}
MacKay, D. J.~C. 2003.
\newblock \emph{Information Theory, Inference and Learning Algorithms}.
\newblock Cambridge University Press.

\bibitem[{{OpenAI}(2026)}]{openai2026gpt54}
{OpenAI}. 2026.
\newblock Introducing {GPT-5.4}.
\newblock Published March 5, 2026. Accessed: 2026-04-28.

\bibitem[{{OpenRouter}(2026{\natexlab{a}})}]{openrouter2026datacollection}
{OpenRouter}. 2026{\natexlab{a}}.
\newblock Data Collection.
\newblock Online documentation.
\newblock Accessed 2026-07-26.

\bibitem[{{OpenRouter}(2026{\natexlab{b}})}]{openrouter2026api}
{OpenRouter}. 2026{\natexlab{b}}.
\newblock OpenRouter API Reference.
\newblock Online documentation.
\newblock Accessed 2026-07-22.

\bibitem[{{OpenRouter}(2026{\natexlab{c}})}]{openrouter2026routing}
{OpenRouter}. 2026{\natexlab{c}}.
\newblock Provider Routing.
\newblock Online documentation.
\newblock Accessed 2026-07-22.

\bibitem[{{OpenRouter}(2026{\natexlab{d}})}]{openrouter2026zdr}
{OpenRouter}. 2026{\natexlab{d}}.
\newblock Zero Data Retention.
\newblock Online documentation.
\newblock Accessed 2026-07-26.

\bibitem[{{Qwen Team}(2026)}]{qwen35blog}
{Qwen Team}. 2026.
\newblock Qwen3.5: Accelerating Productivity with Native Multimodal Agents.

\bibitem[{Sakkalis, Shen, and Patrikalakis(2000)}]{sakkalis2000validity}
Sakkalis, T.; Shen, G.; and Patrikalakis, N.~M. 2000.
\newblock Representational Validity of Boundary Representation Models.
\newblock \emph{Computer-Aided Design}, 32(12): 719--726.

\bibitem[{Snell et~al.(2025)Snell, Lee, Xu, and
  Kumar}]{snell2024scalingllmtesttimecompute}
Snell, C.; Lee, J.; Xu, K.; and Kumar, A. 2025.
\newblock Scaling {LLM} Test-Time Compute Optimally can be More Effective than
  Scaling Model Parameters.
\newblock In \emph{International Conference on Learning Representations}.

\bibitem[{Song, Meng, and Ermon(2021)}]{song2021ddim}
Song, J.; Meng, C.; and Ermon, S. 2021.
\newblock Denoising Diffusion Implicit Models.
\newblock In \emph{International Conference on Learning Representations}.

\bibitem[{Xu et~al.(2024)Xu, Lambourne, Jayaraman, Wang, Willis, and
  Furukawa}]{xu2024brepgen}
Xu, X.; Lambourne, J.; Jayaraman, P.; Wang, Z.; Willis, K.; and Furukawa, Y.
  2024.
\newblock {BrepGen}: A {B}-rep Generative Diffusion Model with Structured
  Latent Geometry.
\newblock \emph{ACM Trans. Graph.}, 43(4).

\bibitem[{Ye et~al.(2024)Ye, Lin, Han, Xu, Liu, Liang, Ma, Zou, and
  Ermon}]{ye2024tfg}
Ye, H.; Lin, H.; Han, J.; Xu, M.; Liu, S.; Liang, Y.; Ma, J.; Zou, J.~Y.; and
  Ermon, S. 2024.
\newblock {TFG}: Unified Training-Free Guidance for Diffusion Models.
\newblock In \emph{Advances in Neural Information Processing Systems},
  volume~37, 22370--22417.

\bibitem[{Yu, Schumacher, and Crane(2021{\natexlab{a}})}]{yu2021repulsive}
Yu, C.; Schumacher, H.; and Crane, K. 2021{\natexlab{a}}.
\newblock Repulsive Curves.
\newblock \emph{ACM Trans. Graph.}, 40(2).

\bibitem[{Yu, Schumacher, and Crane(2021{\natexlab{b}})}]{yu2021repulsivecode}
Yu, C.; Schumacher, H.; and Crane, K. 2021{\natexlab{b}}.
\newblock Repulsive Curves: Reference Implementation.
\newblock GitHub repository.
\newblock Commit 78a0c15da6edfc19a3f6607aebc59d769b1e59da; accessed 2026-07-22.

\end{thebibliography}
\end{document}


\maketitle
\appendix


\section{Analysis on GTAD VLM-based Detector}
\label{sec:gtad-vlm-detector}
The VLM detector provides a coarse visual signal for downstream invalidity risk that complements the geometric and topological detectors. Tab.~\ref{tab:gtad-backbone} keeps the prompt, development cohort, rendering protocol, and evaluation procedure fixed while varying only the VLM backbone~\cite{openai2026gpt54,google2026gemini3flash,anthropic2025claudehaiku45,qwen35blog}. Qwen3.5-Flash attains the highest development-cohort F1-score for predicting whether the downstream B-Rep fails the validity checker and is therefore selected as the VLM coarse-screening backbone before held-out evaluation.

\begin{figure}[t]
  \centering
  \includegraphics[width=\linewidth]{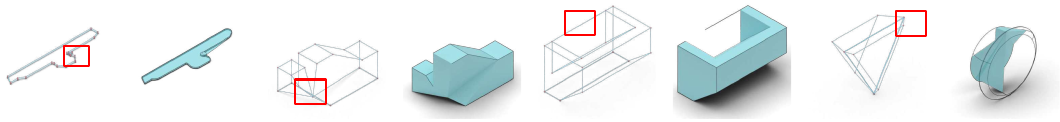}
  \caption{Complementary VLM and topological risk signals. The VLM flags suspicious regions (red boxes) when topology-side checks are negative; each wireframe is paired with its downstream B-Rep.}
  \label{fig:vlm-topo-complementarity}
\end{figure}

\begin{table}[t]
  \centering
  \wdrtablestyle
  \begin{tabular*}{\columnwidth}{@{\extracolsep{\fill}}lcc@{}}
    \toprule
    VLM & F1(\%)$\uparrow$ & Latency(ms)$\downarrow$ \\
    \midrule
    gemini-3-flash & 68.68 & \textbf{1153.4} \\
    claude-haiku-4.5 & 63.58 & 1567.2 \\
    qwen-3.5-flash & \textbf{71.85} & 1678.1 \\
    gpt-5.4-mini & 68.06 & 2136.3 \\
    \bottomrule
  \end{tabular*}
  \caption{VLM invalidity-risk screening on ABC development data. Mean latency includes multi-view rendering, OpenRouter inference~\cite{openrouter2026api}, parsing, and retries.}
  \label{tab:gtad-backbone}
\end{table}
The VLM and topological detectors provide different risk cues rather than duplicating one another: the VLM supplies a broad visual screening signal, whereas the topological detector deterministically evaluates incidence, loop, face, and shell checks. As shown in Fig.~\ref{fig:vlm-topo-complementarity}, the VLM can flag visually suspicious configurations even when the four topology-side checks return negative, motivating their parallel use in GTAD. The VLM output is only a binary routing trigger; it neither identifies a specific anomaly type nor supplies a scalar repair energy.

\section{Analysis of the GTAD Geometric Detector}
\label{sec:gtad-geometric-detector}
\begin{figure}[t]
  \centering
  \includegraphics[width=\linewidth]{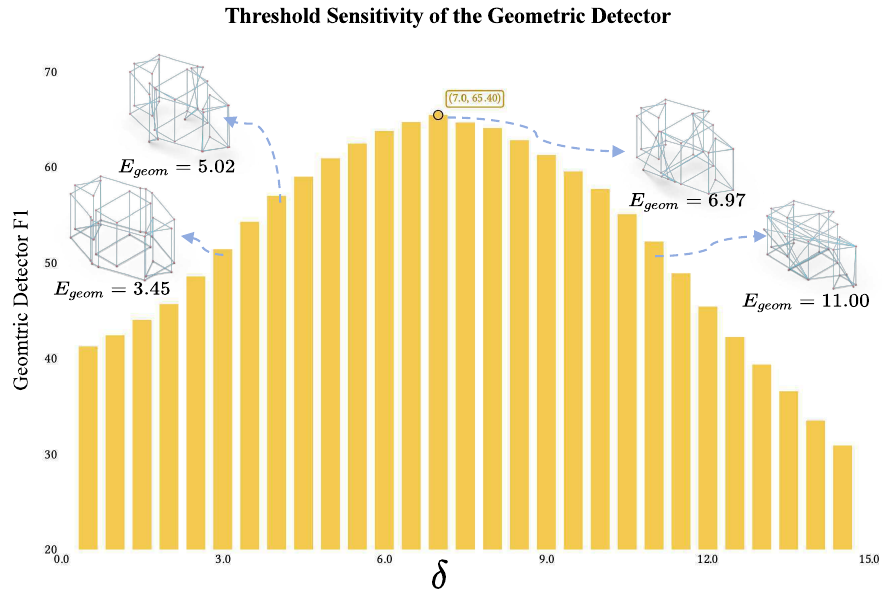}
  \caption{Development-cohort threshold sensitivity of the geometric detector for predicting downstream B-Rep invalidity risk. Sweeping $\delta$ for $E_{\mathrm{geom}}$ yields the best F1 near $\delta=7.0$; the examples show representative candidates at different energy levels.}
  \label{fig:e_geom}
\end{figure}

Fig.~\ref{fig:e_geom} analyzes how the threshold $\delta$ affects the standalone GTAD geometric detector. We sweep $\delta$ on the development cohort and report F1 against the downstream B-Rep invalidity label. The performance first improves as overly strict thresholds are relaxed, reaches the best value around $\delta=7.0$, and then decreases as fewer high-energy candidates are routed. A small threshold tends to over-route benign wireframes, whereas a large threshold may miss high-risk configurations. We therefore fix $\delta=7.0$ before held-out evaluation and use it in all experiments.

\begin{table*}[t]
  \centering
  \wdrtablestyle
  \begin{tabular*}{\textwidth}{@{}l@{\extracolsep{\fill}}clccc@{}}
    \toprule
    Case family & \makecell{Expected\\label} & \makecell{Controlled\\parameter} & Samples &
    \makecell{Mean/median\\$E_{\mathrm{geom}}$} &
    \makecell{Flagged\\samples} \\
    \midrule
    \makecell[l]{Long-segment\\interior crossing} & Invalid & \makecell[l]{Crossing\\angle/offset} & 1,000 & 7.47/7.89 & 795 \\
    \midrule
    \makecell[l]{Near-collapsed\\edge} & Invalid & \makecell[l]{Normalized\\edge length} & 1,000 & 8.23/9.10 & 832 \\
    \midrule
    Legal near-contact & Valid & \makecell[l]{Minimum\\separation} & 1,000 & 2.18/3.54 & 67 \\
    \bottomrule
  \end{tabular*}
  \caption{Controlled validation of the discrete geometric risk score at $\delta=7.0$. The expected label records whether the constructed case is invalid or valid.}
  \label{tab:supp-controlled-geometry}
\end{table*}

\subsection{High-Risk Geometric Candidates and Near-Contact Analysis}
\begin{figure*}[t]
  \centering
  \includegraphics[width=\linewidth]{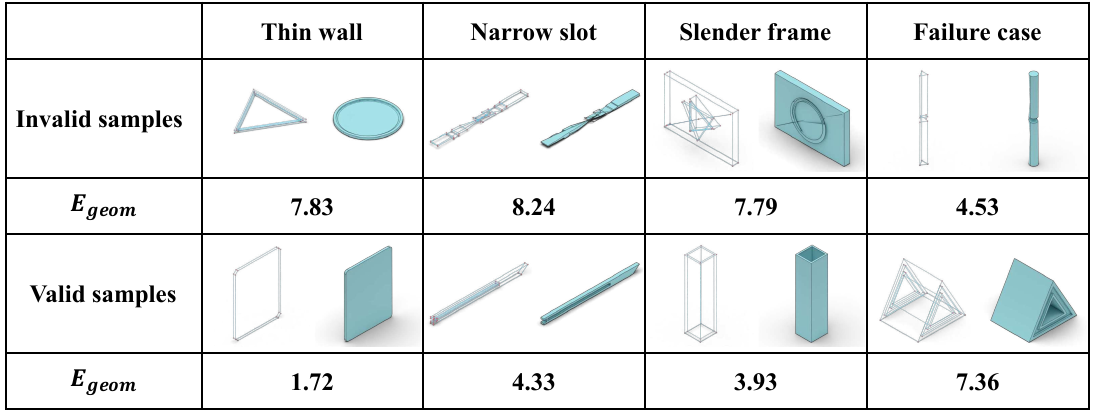}
  \caption{Near-contact analysis of the geometric detector. Representative invalid and valid wireframe/B-Rep pairs cover thin walls, narrow slots, and slender frames. The final column shows counterexamples in which a low-energy sample is invalid and a high-energy sample is valid, illustrating why $E_{\mathrm{geom}}$ is used as a soft risk indicator rather than a standalone validity test.}
  \label{fig:near-contact}
\end{figure*}

Valid CAD shapes may contain thin, narrow, or slender components whose curves are spatially close without being self-intersecting. We therefore further evaluate the geometric detector on an independent cohort of 2,682 valid ABC samples exhibiting such structures. The mean energy over the complete cohort is $E_{\mathrm{geom}}=2.18$, below the selected threshold, while 7.16\% of the samples are flagged as high-risk candidates by $E_{\mathrm{geom}}>7.0$. The low energies of the valid thin-wall, narrow-slot, and slender-frame examples in Fig.~\ref{fig:near-contact} further indicate that close spacing alone is not heavily penalized. These results support treating tangent-point energy as a soft geometry-side risk indicator, rather than as a standalone invalidity test for every near-contact configuration.
Tab.~\ref{tab:supp-controlled-geometry} shows that the detector flags 795 of 1,000 interior-crossing cases and 832 of 1,000 near-collapsed cases, while flagging only 67 of 1,000 legal near-contact cases. The remaining misses and false positives further support using $E_{\mathrm{geom}}$ as a heuristic risk score rather than a complete validity predicate.

\section{Analysis of the GTAD Topology Detector}
\label{sec:gtad-topology-detector}

\paragraph{Topology Constraints.}
For closed, manifold B-Rep models, we evaluate $\mathcal{W}$ using four validity-inspired structural risk checks grounded in standard B-Rep incidence and connectivity definitions. They are calibrated to the native representations of the evaluated generators and serve as routing heuristics rather than universal validity theorems. For each check, the indicator is $1$ when the corresponding risk pattern is detected and $0$ otherwise.
\begin{itemize}

\item \textit{Vertex-Edge Constraint ($C_1$):} In the evaluated wireframe representations, degree below two indicates dangling connectivity, while degree above eight is treated as an unusually complex, high-risk pole~\cite{sakkalis2000validity}. We therefore use $[2,8]$ as a representation-calibrated routing range, not as a universal admissibility condition for every legal B-Rep. We define
\begin{equation}
r_1(\mathcal{W})= \frac{1}{|\mathcal{V}|} \sum_{v\in\mathcal{V}}C_1(v),
\end{equation}
where $C_1(v)=1$ when the degree of vertex $v$ lies outside this calibrated range and $0$ otherwise.

\item \textit{Edge-Loop Constraint ($C_2$):} The evaluated serializers represent ordinary face boundaries as simple cycles containing at least three vertices and no repeated edges~\cite{li2025stitchashape,jayaraman2023solidgen}. We flag departures from this representation convention as routing risks; the minimum loop size is not asserted as a necessary condition for every legal kernel representation. We define
\begin{equation}
r_2(\mathcal{W})= \frac{1}{|\mathcal{L}|} \sum_{l\in\mathcal{L}}C_2(l),
\end{equation}
where $\mathcal{L}$ is the set of boundary loops and $C_2(l)=1$ when loop $l$ violates either condition.

\item \textit{Loop-Face Constraint ($C_3$):} Each face must be trimmed by exactly one outer loop and zero or more inner loops (holes). These closed, oriented loops bound the valid, visible region of the underlying parametric surface~\cite{li2025stitchashape,jayaraman2023solidgen}. We define
\begin{equation}
r_3(\mathcal{W})= \frac{1}{|\mathcal{F}|} \sum_{f\in\mathcal{F}}C_3(f),
\end{equation}
where $\mathcal{F}$ is the set of faces and $C_3(f)=1$ when the trimming-loop configuration of face $f$ violates this constraint.

\item \textit{Face-Shell Constraint ($C_4$):} For an object composed of a single solid shell, its edge-face adjacency graph must form one connected component~\cite{ansaldi1985geometric}. We define
\begin{equation}
r_4(\mathcal{W})= C_4(\mathcal{W}),
\end{equation}
where $C_4(\mathcal{W})=1$ when the global connectivity constraint is violated and $0$ otherwise.

\end{itemize}

The combined topological energy is
\begin{equation}
E_{\mathrm{topo}}(\mathcal{W})=\sum_{i=1}^4 r_i(\mathcal{W}),
\end{equation}
and $E_{\mathrm{topo}}(\mathcal{W})>0$ activates topology-side guided regeneration.

\paragraph{Detector Analysis.}
The topology criteria and their combined routing rule are fixed before held-out evaluation. Tab.~\ref{tab:gtad-topo} evaluates them using F1-score against the downstream B-Rep invalidity label. Each individual criterion provides a distinct structural risk cue, with $C_1$ achieving the strongest individual result. Combining the complementary violations through $E_{\mathrm{topo}}(\wire)>0$ yields the highest F1-score.

\begin{table}[t]
  \centering
  \small
  \begin{tabular*}{\columnwidth}{@{}l@{\extracolsep{\fill}}c@{}}
    \toprule
    Criterion & F1(\%)$\uparrow$ \\
    \midrule
    $C_1$    & 65.32  \\
    $C_2$    & 45.30  \\
    $C_3$    & 24.66  \\
    $C_4$    & 54.62 \\
    \midrule
    $E_{\mathrm{topo}}(\wire)>0$ & \textbf{74.75}  \\
    \bottomrule
  \end{tabular*}
  \caption{Topology-side prediction of downstream B-Rep invalidity risk on the held-out detector test cohort. Criteria are evaluated on pre-guidance topology outputs, with final B-Rep checker failure used as the positive label.}
  \label{tab:gtad-topo}
\end{table}

\begin{figure}[t]
  \centering
  \includegraphics[width=\linewidth]{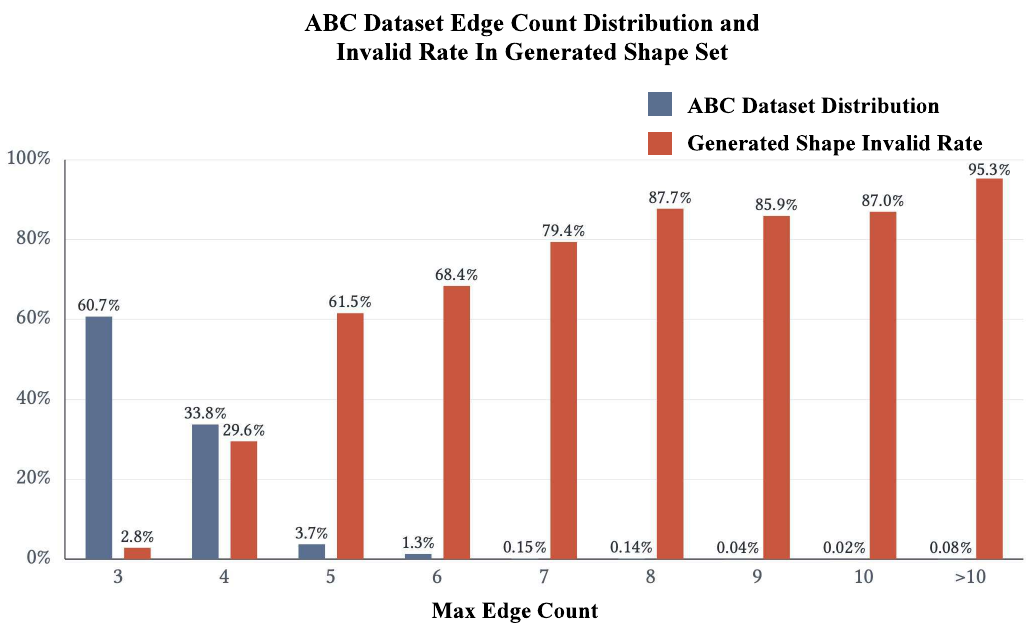}
\caption{Analysis of the distribution of the maximum vertex edge count in the ABC dataset and its relation to invalidity in generated shapes.}

  \label{fig:edge_count}
\end{figure}

\paragraph{Analysis of the Vertex-Edge Constraint.}
We use the \textit{Vertex-Edge Constraint} ($C_1$) to flag dangling vertices and abnormally complex poles. Fig.~\ref{fig:edge_count} shows that, in the ABC dataset, the maximum vertex edge count is concentrated at small values, whereas in generated shapes the invalid rate rises quickly as this count increases. In particular, unusually high-degree vertices are rare in the reference dataset but frequently associated with invalid outputs. This empirical association supports using vertex degree as a representation-calibrated topological risk heuristic in GTAD and motivates the practical upper bound in $C_1$.

\section{GTAD Routing}
\label{sec:supp_gtad_routing}

This section summarizes how the Geometric-Topology Anomaly Detector (GTAD) routes an intermediate wireframe to the Energy-Guided Geometric-Topology Repair (EGGTR) module, as shown in Algorithm~\ref{alg:gtad_routing}. Given the topology-aware wireframe $\mathcal{W}$ later formalized in Eq.~\eqref{eq:supp-topology-aware-wireframe}, GTAD evaluates complementary visual, geometric, and topological signals for downstream invalidity risk in parallel. The routing output is an active branch set $\mathcal{A}_{\mathrm{rep}}\subseteq\{\mathsf{Geom},\mathsf{Topo}\}$, where $\mathsf{Geom}$ requests geometry-guided regeneration or reranking and $\mathsf{Topo}$ requests its topology-guided counterpart. An empty set $\mathcal{A}_{\mathrm{rep}}=\emptyset$ indicates that all three detectors return negative and the original trajectory proceeds without an EGGTR intervention.

Let $D_{\mathrm{V}}(\mathcal W)=1$ denote a VLM output of \texttt{fail}. The three parallel detector outputs are
\begin{equation}
\begin{aligned}
D_{\mathrm{G}}(\mathcal W)
  &=\mathbf{1}\!\left[E_{\mathrm{geom}}(\mathcal W)>\delta\right],\\
D_{\mathrm{T}}(\mathcal W)
  &=\mathbf{1}\!\left[E_{\mathrm{topo}}(\mathcal W)>0\right],\\
\mathbf D_{\mathrm{GTAD}}(\mathcal W)
  &=\bigl[D_{\mathrm{V}}(\mathcal W),D_{\mathrm{G}}(\mathcal W),D_{\mathrm{T}}(\mathcal W)\bigr]^{\mathsf T}.
\end{aligned}
\label{eq:supp-gtad-decision}
\end{equation}
The VLM produces a broad binary screening signal without attributing an anomaly type or defining a scalar energy, whereas the deterministic detectors provide branch-specific signals. We therefore define
\begin{equation}
\begin{aligned}
m_{\mathrm{G}}(\mathcal W)&=\max\!\left(D_{\mathrm{V}}(\mathcal W),D_{\mathrm{G}}(\mathcal W)\right),\\
m_{\mathrm{T}}(\mathcal W)&=\max\!\left(D_{\mathrm{V}}(\mathcal W),D_{\mathrm{T}}(\mathcal W)\right),\\
\mathcal A_{\mathrm{rep}}(\mathcal W)
&=\{\mathsf{Geom}\mid m_{\mathrm{G}}=1\}
  \cup\{\mathsf{Topo}\mid m_{\mathrm{T}}=1\}.
\end{aligned}
\label{eq:supp-gtad-routing-mask}
\end{equation}
Thus, a positive VLM signal activates both guided branches, while positive geometric and topological signals activate their corresponding branches.

\begin{algorithm}[t]
\caption{GTAD Routing}
\label{alg:gtad_routing}
\begin{algorithmic}[1]
\Require Topology-aware wireframe $\mathcal{W}$, VLM detector $R$, geometry threshold $\delta$
\Ensure Active guidance branches $\mathcal{A}_{\mathrm{rep}}\subseteq\{\mathsf{Geom},\mathsf{Topo}\}$

\State $\mathcal{A}_{\mathrm{rep}} \gets \emptyset$
\State $r_{\mathrm{vis}} \gets R(\mathcal{W})$
\Comment{visual risk signal}
\State $e_{\mathrm{geom}} \gets E_{\mathrm{geom}}(\mathcal{W})$
\State $e_{\mathrm{topo}} \gets E_{\mathrm{topo}}(\mathcal{W})$
\Comment{independent detector evaluation}

\If{$r_{\mathrm{vis}} = 1$}
    \State $\mathcal{A}_{\mathrm{rep}} \gets \mathcal{A}_{\mathrm{rep}} \cup \{\mathsf{Geom},\mathsf{Topo}\}$
\EndIf

\If{$e_{\mathrm{geom}} > \delta$}
    \State $\mathcal{A}_{\mathrm{rep}} \gets \mathcal{A}_{\mathrm{rep}} \cup \{\mathsf{Geom}\}$
\EndIf

\If{$e_{\mathrm{topo}} > 0$}
    \State $\mathcal{A}_{\mathrm{rep}} \gets \mathcal{A}_{\mathrm{rep}} \cup \{\mathsf{Topo}\}$
\EndIf

\State \Return $\mathcal{A}_{\mathrm{rep}}$
\end{algorithmic}
\end{algorithm}

GTAD only determines whether and how the current generation trajectory should be revisited; it does not update the generator parameters. Here, ``repair'' denotes detector-triggered guided regeneration or candidate reranking from an exposed intermediate representation, rather than deterministic geometric editing of a fixed, completed wireframe. The three detectors operate in parallel, and each independently contributes to routing. A geometric or topological detection activates its corresponding branch. Because the VLM signal has no explicit anomaly type, a positive VLM decision activates both branches. Multiple positive signals jointly determine the active branches, and the original trajectory proceeds unchanged only when all three detectors return negative. Tab.~\ref{tab:supp-routing-e2e} summarizes the corresponding end-to-end routing ablation.

\begin{table*}[t]
  \centering
  \wdrtablestyle
  \begin{tabular*}{\textwidth}{@{}l@{\extracolsep{\fill}}cccc@{}}
    \toprule
    Routing &
    \makecell{Final\\Valid(\%)$\uparrow$} &
    \makecell{Invalid\\$\rightarrow$Valid$\uparrow$} &
    \makecell{Valid\\$\rightarrow$Invalid$\downarrow$} &
    \makecell{Latency\\(ms/sample)$\downarrow$} \\
    \midrule
    Unguided baseline & 68.1 & 0 & 0 & 683 \\
    VLM-only & 77.4 & 415 & 136 & 2457 \\
    Geom. + Topo. (w/o VLM) & 78.6 & 452 & 137 & 836 \\
    Full & \textbf{83.9} & 553 & 79 & 3351 \\
    \bottomrule
  \end{tabular*}
  \caption{End-to-end routing ablation on DTGBrepGen and ABC using 3,000 paired samples. Invalid$\rightarrow$Valid and Valid$\rightarrow$Invalid are counts relative to the unguided output; latency is measured end to end in milliseconds per sample. Full denotes parallel VLM, Geom., and Topo. routing with the default resampling configuration.}
  \label{tab:supp-routing-e2e}
\end{table*}
The full configuration achieves the highest \emph{Valid} score, improving the unguided baseline by 15.8 points while converting 553 initially invalid outputs to valid and 79 initially valid outputs to invalid. Relative to VLM-only and Geom.+Topo., it improves \emph{Valid} by 6.5 and 5.3 points, respectively, at the cost of higher latency.

To separate detector-informed routing from uniformly increasing test-time computation, Tab.~\ref{tab:supp-routing-controls} compares the current GTAD configuration with two controls under the same default EGGTR configuration, condition IDs, and base random seeds. \emph{Always-on EGGTR} activates both guidance branches for every sample. \emph{Matched random routing} preserves the numbers of geometry-only, topology-only, and joint branch activations produced by the current configuration, but assigns them uniformly at random using a fixed routing seed.
\begin{table*}[t]
  \centering
  \wdrtablestyle
  \begin{tabular*}{\textwidth}{@{}l@{\extracolsep{\fill}}cccccc@{}}
    \toprule
    Routing &
    \makecell{Valid\\(\%)$\uparrow$} &
    \makecell{$I\!\rightarrow\!V$\\$\uparrow$} &
    \makecell{$V\!\rightarrow\!I$\\$\downarrow$} &
    \makecell{Novel\\(\%)$\uparrow$} &
    \makecell{Unique\\(\%)$\uparrow$} &
    \makecell{Latency\\(ms/sample)$\downarrow$} \\
    \midrule
    \makecell[l]{Current GTAD\\routing}
      & 83.9 & 553 & 79 & 99.9 & 99.3 & 3351 \\
    \midrule
    \makecell[l]{Always-on\\EGGTR}
      & 84.3 & 697 & 211 & 99.7 & 99.1 & 1354 \\
    \midrule
    \makecell[l]{Matched random\\routing}
      & 80.2 & 536 & 174 & 99.8 & 98.6 & 1067 \\
    \bottomrule
  \end{tabular*}
  \caption{End-to-end routing controls on DTGBrepGen and ABC using 3,000 paired samples. $I\!\rightarrow\!V$ and $V\!\rightarrow\!I$ are counts relative to the unguided outputs; latency is measured end to end in milliseconds per sample.}
  \label{tab:supp-routing-controls}
\end{table*}
\paragraph{Why GTAD instead of always-on EGGTR?}
Always-on EGGTR obtains the highest raw \emph{Valid} score, but its 0.4-point advantage over the current GTAD routing corresponds to only 12 additional net-valid outputs: it converts 144 more invalid outputs to valid while also converting 132 more initially valid outputs to invalid. It also reduces \emph{Novel} and \emph{Unique} by 0.2 points each. In contrast, matched random routing uses the same branch-activation counts as GTAD but reaches only 80.2\% \emph{Valid}. It recovers 17 fewer invalid samples and introduces 95 more Valid$\rightarrow$Invalid regressions than the detector-informed routing, while reducing \emph{Unique} by 0.7 points. This comparison isolates the benefit of assigning interventions to detector-identified samples rather than matching intervention volume alone. The lower latency of the two controls reflects their omission of detector evaluation, including remote VLM inference. Overall, GTAD routing retains nearly the raw validity of always-on guidance while substantially reducing harmful interventions and preserving stronger diversity, at the cost of additional screening latency.

\section{Comparison of Sampling Strategies}
\label{sec:supp-sampling-strategies}

\begin{figure*}[!t]
  \centering
  \includegraphics[width=0.85\linewidth]{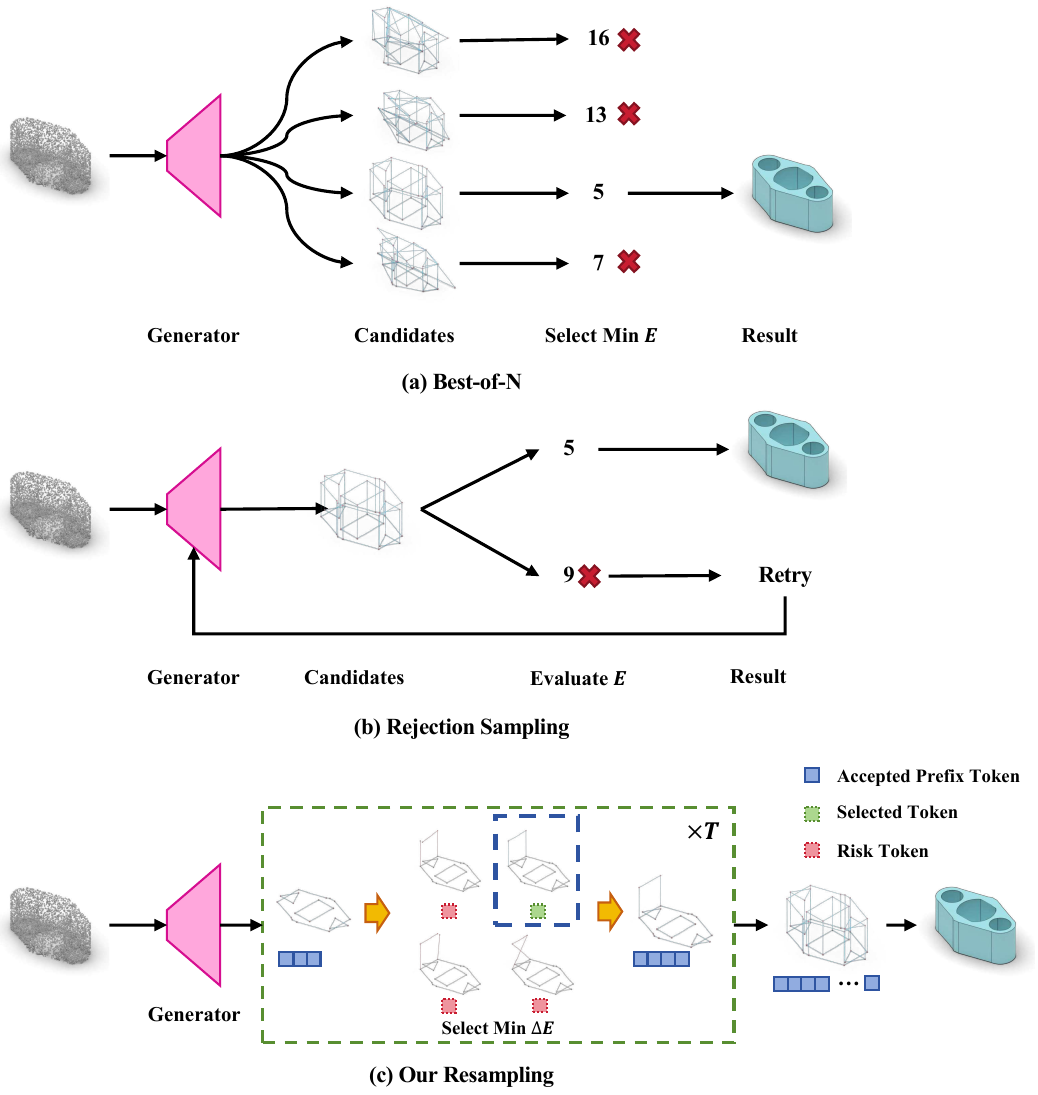}
  \caption{Conceptual comparison of the three sampling strategies at the default $K=N=4$ setting. The one-retry rejection-sampling variant discards a flagged complete trajectory and allows one full retry; Best-of-$N$ generates four complete candidates in parallel before selecting the lowest-energy result; our resampling evaluates four token candidates on the current partial wireframe and preserves the accepted prefix.}
  \label{fig:sampling-strategies}
\end{figure*}

We compare inference-time strategies under the same condition and base random seed. Tab.~\ref{tab:supp-sampling-budget} reports their default-setting compute profiles. A generator-specific decoder exposes a topology-aware intermediate representation.
Fig.~\ref{fig:sampling-strategies} contrasts where the three strategies allocate additional computation and when candidate selection is performed.

\begin{table*}[t]
  \centering
  \wdrtablestyle
  \begin{tabular*}{\textwidth}{@{}l@{\extracolsep{\fill}}cccccc@{}}
    \toprule
    Method &
    \makecell{Valid\\(\%)$\uparrow$} &
    \makecell{Novel\\(\%)$\uparrow$} &
    \makecell{Unique\\(\%)$\uparrow$} &
    \makecell{Sampling\\(ms/sample)$\downarrow$} &
    \makecell{E2E\\(ms/sample)$\downarrow$} &
    \makecell{Peak mem.\\(GB)$\downarrow$} \\
    \midrule
    Baseline & 68.1 & \textbf{99.9} & \textbf{99.4} & \textbf{654} & \textbf{683} & \textbf{31.5} \\
    \midrule
    \makecell[l]{Geometry + Topology\\(GTAD-triggered\\One-Retry)}
      & 73.6 & 99.1 & 97.9 & 1177 & 2941 & 32.4 \\
    \midrule
    \makecell[l]{OCCT-routed\\One-Retry}
      & 78.2 & 98.9 & 97.6 & 1340 & 3842 & 31.6 \\
    \midrule
    \makecell[l]{Geometry + Topology\\(Best-of-$4$)}
      & 80.7 & 99.2 & 98.6 & 878 & 2735 & 67.9 \\
    \midrule
    Our Resampling & \textbf{83.9} & \textbf{99.9} & 99.3 & 1489 & 3351 & 33.5 \\
    \bottomrule
  \end{tabular*}
  \caption{Compute-profile comparison of sampling strategies. The GTAD-triggered and OCCT-triggered variants use the same maximum-one-retry budget, but the latter receives feedback only after final B-Rep construction. Best-of-$N$ generates its complete trajectories in parallel as one batch. All time values are reported in milliseconds per sample. Sampling wall-clock time excludes detector or checker feedback, rendering, and final B-Rep construction. End-to-end latency covers each method's complete path from initial generation to the returned B-Rep, including method-specific feedback, retry or reranking, and final construction. Peak memory is the maximum allocated GPU memory on an NVIDIA A100. Bold indicates the best result in each column; ties are bolded.}
  \label{tab:supp-sampling-budget}
\end{table*}
Compared with the OCCT-triggered one-retry baseline, our resampling improves \emph{Valid}, \emph{Novel}, and \emph{Unique} by 5.7, 1.0, and 1.7 points, respectively, while reducing end-to-end latency by 491 ms/sample; peak memory increases moderately by 1.9 GB. Relative to Best-of-$4$, our method improves \emph{Valid} by 3.2 points and reduces peak memory by 34.4 GB, although parallel Best-of-$4$ remains 616 ms/sample faster. These results indicate a favorable validity--diversity--resource trade-off rather than uniform dominance in every cost metric.

\begin{equation}
\begin{aligned}
\mathcal W&=(\mathcal V,\mathcal E,\mathcal T),\\
\mathcal T&=
\left(
\begin{gathered}
\mathcal L,\mathcal F,\mathcal S,
\mathcal R_{\mathrm{VE}},\mathcal R_{\mathrm{EL}},\\
\mathcal R_{\mathrm{LF}},\mathcal R_{\mathrm{FS}}
\end{gathered}
\right),
\end{aligned}
\label{eq:supp-topology-aware-wireframe}
\end{equation}
where $\mathcal V$ and $\mathcal E$ are vertices and edges, $\mathcal L$, $\mathcal F$, and $\mathcal S$ are loops, faces, and shells, and the four relations encode vertex--edge, edge--loop, loop--face, and face--shell incidence. Loop membership in $\mathcal R_{\mathrm{EL}}$ retains boundary order and orientation.

Let $\mathbf x=(x_1,\ldots,x_T)$ denote the autoregressive token sequence under condition $c$. The base generator factorizes as
\begin{equation}
f_\phi(\mathbf x\mid c)
=\prod_{t=1}^{T}
f_\phi(x_t\mid\mathbf x_{<t},c).
\label{eq:supp-ar-factorization}
\end{equation}
At step $t$, the generator-specific prefix decoder exposes the partial representation
\begin{equation}
\mathcal W_t
=\mathcal D_t(\mathbf x_{\leq t}).
\label{eq:supp-prefix-decoder}
\end{equation}
For branch $r\in\{\mathsf{Geom},\mathsf{Topo}\}$, let $\mathcal M_r^{(t)}$ denote the primitives and incidence relations that are complete enough to evaluate at step $t$. We use the masked prefix energy
\begin{equation}
\begin{aligned}
E_r^{(t)}(\mathcal W_t)
&=E_r\!\left(\mathcal W_t;\mathcal M_r^{(t)}\right),\\
E_{\mathcal A}^{(t)}(\mathcal W_t)
&=\sum_{r\in\mathcal A}\lambda_r E_r^{(t)}(\mathcal W_t),
\end{aligned}
\label{eq:supp-masked-prefix-energy}
\end{equation}
where $\mathcal W_{\mathrm{trigger}}$ is the detected intermediate wireframe that triggers the current EGGTR call, $\mathcal A=\mathcal A_{\mathrm{rep}}(\mathcal W_{\mathrm{trigger}})$ is fixed during that call, and $\lambda_r>0$ weights active branch $r$; incomplete primitives or relations are masked. The VLM contributes a binary routing trigger but no scalar energy term: when it returns \texttt{fail}, $\mathcal A$ contains both branches and Eq.~\eqref{eq:supp-masked-prefix-energy} combines their deterministic energies. For a completed candidate,
\begin{equation}
S_{\mathcal A}(\mathcal W)
=E_{\mathcal A}^{(T)}(\mathcal W).
\label{eq:supp-complete-score}
\end{equation}

An energy-tilted sequence distribution provides the conceptual motivation,
\begin{equation}
q^\star(\mathbf x\mid c,\mathcal A)
=\frac{1}{Z(c,\mathcal A)}
f_\phi(\mathbf x\mid c)
\exp\!\left[
-E_{\mathcal A}^{(T)}
\bigl(\mathcal D_T(\mathbf x)\bigr)
\right],
\label{eq:supp-global-energy-tilt}
\end{equation}

\paragraph{Rejection Sampling.}
Our detector-based full-trajectory baseline is a one-retry variant motivated by the classical rejection-sampling principle~\cite{mackay2003information}. Define the scalar trigger
\begin{equation}
\begin{aligned}
D_{\mathrm{any}}(\mathcal W)
&=\max_{r\in\{\mathrm V,\mathrm G,\mathrm T\}}D_r(\mathcal W)\\
&=\mathbf 1\!\left[
\mathcal A_{\mathrm{rep}}(\mathcal W)\neq\emptyset
\right].
\end{aligned}
\label{eq:supp-any-trigger}
\end{equation}
The baseline evaluates a complete wireframe and regenerates the full trajectory once when any GTAD branch is activated. With one initial sample $\mathcal W^{(0)}$ and at most one retry $\mathcal W^{(1)}$, the returned sample is
\begin{equation}
\mathcal W_{\mathrm{RS}}=
\begin{cases}
\mathcal W^{(0)}, & D_{\mathrm{any}}(\mathcal W^{(0)})=0,\\
\mathcal W^{(1)}, & D_{\mathrm{any}}(\mathcal W^{(0)})=1.
\end{cases}
\label{eq:supp-rejection}
\end{equation}
The retry is retained even if it is also flagged; a subsequent B-Rep construction failure is counted as invalid. Thus, this baseline is not unrestricted rejection sampling and performs no deterministic editing of the first complete wireframe.

\paragraph{Best-of-$N$.}
Best-of-$N$ performs complete-trajectory generation followed by test-time reranking~\cite{collins_koo_2005_discriminative,snell2024scalingllmtesttimecompute}: it generates the $N$ complete wireframes in parallel as a single GPU batch and selects the candidate with the lowest branch energy,
\begin{equation}
n^*=\arg\min_{1\leq n\leq N}S_{\mathcal A}(\mathcal W^{(n)}),
\qquad
\mathcal W_{\mathrm{BoN}}=\mathcal W^{(n^*)}.
\label{eq:supp-best-of-n}
\end{equation}
Ties are resolved by retaining the candidate with the smallest generation index. Unlike rejection sampling, this strategy always evaluates $N$ complete decoding trajectories before making a decision; the default comparison uses $N=4$. Its reported latency is batched wall-clock time rather than the sum of per-trajectory runtimes, so the parallel implementation trades lower latency for higher peak GPU memory.

\paragraph{Our Resampling.}
Our method reallocates this additional computation to partial wireframes through local energy-aware candidate reranking. At an evaluable token position $t$, let $\pi_t(\cdot)=f_{\phi,\tau,\operatorname{top-}k}(\cdot\mid\mathbf x_{<t},c)$ denote the temperature-scaled, top-$k$-truncated base proposal and let $K_t=\min(K,|\operatorname{supp}\pi_t|)$. We draw from $\pi_t$ and resample duplicate tokens until obtaining $K_t$ distinct candidates:
\begin{equation}
\begin{aligned}
x_t^{(k)}&\sim\pi_t,\quad k=1,\ldots,K_t,\\
\mathcal C_t&=\{x_t^{(k)}\}_{k=1}^{K_t},
\qquad |\mathcal C_t|=K_t.
\end{aligned}
\label{eq:supp-ar-candidates}
\end{equation}
For each proposed token, the prefix decoder constructs
$\mathcal W_t^{(k)}=\mathcal D_t(\mathbf x_{<t},x_t^{(k)})$.
\paragraph{Shared prefix-energy evaluation.}
For each decoding step $t$, the evaluable mask
$\mathcal M_r^{(t)}$ is determined once from the current decoder state
and token type before the candidate values are instantiated,
and is shared by all candidates in $\mathcal C_t$. Consequently,
the evaluated terms and their normalization denominator remain
fixed within the candidate set. The coordinate normalization
frame is likewise frozen before candidate expansion and reused
for all candidates. Candidate scores therefore differ only in
their geometric or topological values evaluated under the same
context, rather than through candidate-dependent entity counts
or coordinate rescaling.
The corresponding masked energy increment is
\begin{equation}
\Delta E_{\mathcal A}^{(t,k)}
=E_{\mathcal A}^{(t)}(\mathcal W_t^{(k)})
-E_{\mathcal A}^{(t-1)}(\mathcal W_{t-1}).
\label{eq:supp-prefix-increment}
\end{equation}
We form the normalized local surrogate over the sampled candidate indices:
\begin{equation}
\begin{aligned}
w_t^{(k)}
&=\pi_t(x_t^{(k)})
\exp\!\left[-\Delta E_{\mathcal A}^{(t,k)}\right],\\
\widetilde q_t(k\mid\mathcal C_t)
&=\frac{w_t^{(k)}}{\sum_{j=1}^{K_t}w_t^{(j)}}.
\end{aligned}
\label{eq:supp-local-surrogate}
\end{equation}
Within $\mathcal C_t$, the denominator is constant with respect to $k$, so the negative log of Eq.~\eqref{eq:supp-local-surrogate} differs from
\begin{equation}
\ell_t^{(k)}
=-\log \pi_t(x_t^{(k)})
+\Delta E_{\mathcal A}^{(t,k)}
\label{eq:supp-token-resampling}
\end{equation}
only by an additive constant. We therefore apply the deterministic, MAP-like selection rule
\begin{equation}
k^*=\arg\min_{1\leq k\leq K_t}\ell_t^{(k)},
\qquad
x_t\leftarrow x_t^{(k^*)}.
\label{eq:supp-local-candidate-selection}
\end{equation}
We use $K=4$, $\tau=1$, and top-$k=4$ in the default configuration. Positions at which no active energy is evaluable follow a single draw from the original proposal. The first proposal uses the baseline random stream, while additional proposals use deterministic candidate-indexed substreams. Equation~\eqref{eq:supp-local-candidate-selection} is an approximate local selection rule over finite candidates, not exact sampling from the global energy-tilted distribution in Eq.~\eqref{eq:supp-global-energy-tilt}. It concentrates computation where risk becomes measurable while preserving the accepted prefix. For the $K/N$ sweep in Tab.~\ref{tab:supp-sampling-sweep}, we set top-$k=K$ and evaluate $K,N\in\{2,4,8\}$.

\begin{table*}[t]
  \centering
  \wdrtablestyle
  \begin{tabular*}{\textwidth}{@{}l@{\extracolsep{\fill}}ccccc@{}}
    \toprule
    Method &
    \makecell{Valid\\(\%)$\uparrow$} &
    \makecell{Novel\\(\%)$\uparrow$} &
    \makecell{Unique\\(\%)$\uparrow$} &
    \makecell{Latency\\(ms/sample)$\downarrow$} &
    \makecell{Peak\\memory\\(GB)$\downarrow$} \\
    \midrule
    \makecell[l]{Our Resampling\\($K=2$)}
      & \textbf{78.4} & \textbf{99.9} & \textbf{99.2} & 2925 & \textbf{32.3} \\
    \makecell[l]{Our Resampling\\($K=4$)}
      & \textbf{83.9} & \textbf{99.9} & \textbf{99.3} & 3351 & \textbf{33.5} \\
    \makecell[l]{Our Resampling\\($K=8$)}
      & \textbf{84.6} & \textbf{99.9} & \textbf{99.2} & 3875 & \textbf{41.2} \\
    \midrule
    \makecell[l]{Best-of-$N$\\($N=2$)}
      & 77.4 & 99.8 & 99.1 & \textbf{2125} & 45.3 \\
    \makecell[l]{Best-of-$N$\\($N=4$)}
      & 80.7 & 99.2 & 98.6 & \textbf{2735} & 67.9 \\
    \makecell[l]{Best-of-$N$\\($N=8$)}
      & 82.6 & 99.0 & 97.4 & \textbf{3456} & 74.1 \\
    \bottomrule
  \end{tabular*}
  \caption{$K/N$ sweep for local resampling and Best-of-$N$ on DTGBrepGen and ABC. Best-of-$N$ generates its $N$ complete trajectories in parallel as one batch; its latency is therefore batched wall-clock time. Latency is measured end to end in milliseconds per sample, and peak memory is the maximum allocated GPU memory on an NVIDIA A100. Bold indicates the better result within each paired $K=N$ setting.}
  \label{tab:supp-sampling-sweep}
\end{table*}

Local candidates are partial wireframes, whereas Best-of-$N$ candidates are complete trajectories generated in parallel; paired $K=N$ settings therefore need not have equal wall-clock or memory profiles. Across these settings, local resampling improves \emph{Valid} by 1.0--3.2 points and uses 13.0--34.4 GB less peak memory while retaining higher \emph{Novel} and \emph{Unique}; parallel Best-of-$N$ is 419--800 ms/sample faster. Increasing $K$ from 4 to 8 yields only a further 0.7-point \emph{Valid} gain while adding 524 ms/sample and 7.7 GB, motivating $K=4$ as the default trade-off.

\section{Training-Free Guidance Procedure}
\label{sec:supp-complete-tfg}
Training-Free Guidance (TFG) is invoked only when GTAD activates the geometry route of the diffusion-based DTGBrepGen stage; the discrete topology route remains autoregressive. Let $c$ denote the original generator condition and let $p_{\theta}(x_0\mid c)$ be the clean-data distribution induced by the frozen conditional diffusion model. The geometry-energy factor and its normalized target are
\begin{equation}
\begin{aligned}
g_{\lambda}(x_0)
&=\exp\!\left[-\lambda_{\mathrm{geom}}
E_{\mathrm{geom}}\!\left(\mathcal W(x_0)\right)\right],\\
q_{\lambda}(x_0\mid c)
&=\frac{p_{\theta}(x_0\mid c)g_{\lambda}(x_0)}
{Z_{\lambda}(c)},\\
Z_{\lambda}(c)
&=\int p_{\theta}(\tilde x_0\mid c)g_{\lambda}(\tilde x_0)
\,\mathrm d\tilde x_0.
\end{aligned}
\label{eq:supp-tfg-target}
\end{equation}
Thus, the auxiliary energy tilts the conditional base distribution without replacing or dropping $c$.

\paragraph{Forward process and clean estimate.}
The one-step forward noising kernel and cumulative retention are
\begin{equation}
\begin{aligned}
q_{\mathrm{fwd}}(x_t\mid x_{t-1})
&=\mathcal N\!\left(\sqrt{\alpha_t}x_{t-1},
(1-\alpha_t)I\right),\\
\bar\alpha_t&=\prod_{s=1}^{t}\alpha_s.
\end{aligned}
\label{eq:supp-tfg-forward}
\end{equation}
Equivalently, $x_t=\sqrt{\bar\alpha_t}x_0+\sqrt{1-\bar\alpha_t}\epsilon$ with $\epsilon\sim\mathcal N(0,I)$. The frozen conditional noise predictor produces the Tweedie clean estimate
\begin{equation}
\hat x_{0,t}(x_t,c)
=\frac{x_t-\sqrt{1-\bar\alpha_t}\,
\epsilon_{\theta}(x_t,t,c)}
{\sqrt{\bar\alpha_t}}.
\label{eq:supp-tfg-tweedie}
\end{equation}

\paragraph{Smoothed energy guidance.}
Following TFG~\cite{ye2024tfg}, we smooth the energy factor at step $t$ and use a Monte Carlo estimator:
\begin{equation}
\begin{aligned}
\widetilde g_t(u)
&=\mathbb E_{\delta\sim\mathcal N(0,I)}
\!\left[g_{\lambda}(u+\sigma_t\delta)\right],\\
\widehat g_t^{(M)}(u)
&=\frac{1}{M}\sum_{m=1}^{M}
g_{\lambda}(u+\sigma_t\delta_m),\\
\mathcal L_t(u)&=-\log\widehat g_t^{(M)}(u),
\qquad \delta_m\sim\mathcal N(0,I).
\end{aligned}
\label{eq:supp-tfg-smoothing}
\end{equation}
We use the single-sample estimator $M=1$, matching the standard TFG experimental configuration. At recurrence $r$, write $\hat x_{0,t}^{(r)}=\hat x_{0,t}(x_t^{(r)},c)$. The noisy-state guidance and iterated clean-estimate guidance are
\begin{equation}
\begin{aligned}
\Delta_t^{(r)}
&=-\rho_t\nabla_{x_t^{(r)}}
\mathcal L_t\!\left(\hat x_{0,t}(x_t^{(r)},c)\right),\\
\Delta_0^{(r,0)}&=0,\\
\Delta_0^{(r,j+1)}
&=\Delta_0^{(r,j)}
-\mu_t\left.\nabla_u\mathcal L_t(u)\right|_{
u=\hat x_{0,t}^{(r)}+\Delta_0^{(r,j)}} ,
\end{aligned}
\label{eq:supp-tfg-gradients}
\end{equation}
for $j=0,\ldots,N_{\mathrm{iter}}-1$. The first gradient differentiates through Eq.~\eqref{eq:supp-tfg-tweedie}; the second directly updates the clean estimate. The diffusion parameters $\theta$ remain frozen.

\paragraph{Reverse update and recurrence.}
Let $\operatorname{Sample}_{\theta}(\cdot)$ denote the standard conditional DDIM~\cite{song2021ddim} reverse step. Each recurrence first applies
\begin{equation}
\begin{aligned}
x_{t-1}^{(r)}
&=\operatorname{Sample}_{\theta}\!\left(
x_t^{(r)},\hat x_{0,t}^{(r)},c,t\right)\\
&\quad+\frac{\Delta_t^{(r)}}{\sqrt{\alpha_t}}
+\sqrt{\bar\alpha_{t-1}}\,
\Delta_0^{(r,N_{\mathrm{iter}})}.
\end{aligned}
\label{eq:supp-tfg-reverse}
\end{equation}
Set $x_t^{(0)}=x_t$. Only between recurrent reverse updates, for $r=0,\ldots,N_{\mathrm{recur}}-2$, we re-noise with the forward kernel:
\begin{equation}
\begin{aligned}
x_t^{(r+1)}
&\sim q_{\mathrm{fwd}}\!\left(
x_t\mid x_{t-1}^{(r)}\right)\\
&=\mathcal N\!\left(
\sqrt{\alpha_t}x_{t-1}^{(r)},(1-\alpha_t)I\right).
\end{aligned}
\label{eq:supp-tfg-recurrence}
\end{equation}
After the final recurrence, the state passed to the next reverse timestep is $x_{t-1}=x_{t-1}^{(N_{\mathrm{recur}}-1)}$. Thus, Eq.~\eqref{eq:supp-tfg-recurrence} is a recurrence-only forward re-noising operation and is not part of the reverse transition in Eq.~\eqref{eq:supp-tfg-reverse}.

\paragraph{Schedules and reported configuration.}
We process $t=T,T-1,\ldots,1$ using $T=250$ DDIM inference steps and set $N_{\mathrm{iter}}=N_{\mathrm{recur}}=2$. The schedule amplitudes are $\bar\rho=\bar\mu=1$ and $\bar\gamma=0.01$:
\begin{equation}
\begin{aligned}
\rho_t&=\bar\rho
\frac{T\alpha_t}{\sum_{s=1}^{T}\alpha_s},\\
\mu_t&=\bar\mu
\frac{T\alpha_t}{\sum_{s=1}^{T}\alpha_s},\\
\sigma_t&=\bar\gamma\sqrt{1-\bar\alpha_t}.
\end{aligned}
\label{eq:supp-tfg-schedules}
\end{equation}
These yield increasing guidance schedules and a decreasing smoothing scale along the reverse trajectory. We set $\lambda_{\mathrm{geom}}=1$ for diffusion guidance.

\begin{algorithm}[t]
\caption{Conditional Energy-Guided TFG for Geometry Regeneration}
\label{alg:supp-tfg}
\begin{algorithmic}[1]
\Require Condition $c$, frozen model $\epsilon_{\theta}$, geometry energy $E_{\mathrm{geom}}$, schedules $\{\rho_t,\mu_t,\sigma_t\}_{t=1}^{T}$
\Ensure Guided clean geometry sample $x_0$
\State $x_T\sim\mathcal N(0,I)$
\For{$t=T,T-1,\ldots,1$}
  \State $x_t^{(0)}\gets x_t$
  \For{$r=0,\ldots,N_{\mathrm{recur}}-1$}
    \State Compute $\hat x_{0,t}^{(r)}$ by Eq.~\eqref{eq:supp-tfg-tweedie}
    \State Construct $\mathcal L_t$ by Eq.~\eqref{eq:supp-tfg-smoothing}
    \State Compute $\Delta_t^{(r)}$ by Eq.~\eqref{eq:supp-tfg-gradients}
    \State $\Delta_0^{(r,0)}\gets 0$
    \For{$j=0,\ldots,N_{\mathrm{iter}}-1$}
      \State Update $\Delta_0^{(r,j+1)}$ by Eq.~\eqref{eq:supp-tfg-gradients}
    \EndFor
    \State Compute $x_{t-1}^{(r)}$ by Eq.~\eqref{eq:supp-tfg-reverse}
    \If{$r<N_{\mathrm{recur}}-1$}
      \State Re-noise to $x_t^{(r+1)}$ by Eq.~\eqref{eq:supp-tfg-recurrence}
    \EndIf
  \EndFor
  \State $x_{t-1}\gets x_{t-1}^{(N_{\mathrm{recur}}-1)}$
\EndFor
\State \Return $x_0$
\end{algorithmic}
\end{algorithm}

\section{Implementation Details}
\label{sec:supp-implementation-details}

\subsection{Generator Integration and Data Splits}
\methodname is attached only where a generator exposes an intermediate wireframe before surface construction. Tab.~\ref{tab:supp-generator-integration} summarizes the intervention points and the guidance mechanism used by each branch.
\begin{table}[t]
  \centering
  \normalsize
  \begin{tabular*}{\columnwidth}{@{\extracolsep{\fill}}lccc@{}}
    \toprule
    Generator & Intervention point & Topo. & Geom. \\
    \midrule
    DTGBrepGen & \makecell{after VertGeom;\\before EdgeGeom/\\FaceGeom} & AR & \makecell{Diffusion\\(TFG)} \\
    \makecell[l]{Stitch-A-\\Shape} & \makecell{after curve geometry\\and boundary-loop\\identification} & AR & AR \\
    BrepForge & \makecell{after EdgeWireGPT;\\before surface\\instantiation} & AR & AR \\
    \bottomrule
  \end{tabular*}
  \caption{Generator-specific integration of WDR. ``Topo.'' and ``Geom.'' denote topology-side and geometry-side guidance, respectively; AR denotes autoregressive guidance.}
  \label{tab:supp-generator-integration}
\end{table}
WDR does not introduce a new tokenization; it follows the native serialization and primitive-completion rules of DTGBrepGen, Stitch-A-Shape, and BrepForge~\cite{li2025dtgbrepgen,li2025stitchashape,brepforge2026}. Once GTAD activates a branch, WDR revisits the generator-specific stage listed in Tab.~\ref{tab:supp-generator-integration}, retains the original condition and unaffected upstream state, and reruns the affected stage together with its downstream modules. For the diffusion-based geometry branch of DTGBrepGen, the denoised VertGeom prediction is interpreted as the current wireframe and updated using the TFG procedure described in the preceding \textit{Training-Free Guidance Procedure} section~\cite{ye2024tfg}.

The detector benchmark uses disjoint development and held-out test cohorts of intermediate wireframes generated by DTGBrepGen on ABC. The 2,787-sample development cohort is used only to select the VLM backbone and geometric threshold. Qwen3.5-Flash, the fixed prompt, $\delta=7.0$, the topology criteria, and the routing rule are then frozen before evaluation on a separate 2,787-sample held-out test cohort; no sample is shared between the two cohorts, and no detector parameter is trained. An unguided downstream B-Rep that fails the validity checker provides a positive benchmark label, whereas a checker-valid result provides a negative label; runtime branch routing is determined solely by GTAD. Tab.~\ref{tab:supp-gtad-protocol} summarizes this split and evaluation target. For generator-level comparisons, the baseline and WDR variants use the same test identifiers, condition $c$, and base random seed. Diffusion comparisons additionally share the same initial noise, while extra complete candidates use deterministic seed substreams.
\begin{table}[t]
  \centering
  \normalsize
  \begin{tabular*}{\columnwidth}{@{\extracolsep{\fill}}ll@{}}
    \toprule
    Field & Value \\
    \midrule
    Generator & DTGBrepGen \\
    Dataset & ABC \\
    Development cohort &
      \makecell[l]{2,787 samples: 914 positive,\\1,873 negative} \\
    Development usage &
      \makecell[l]{VLM backbone and geometry\\threshold $\delta$} \\
    Held-out test cohort &
      \makecell[l]{2,787 samples: 889 positive,\\1,898 negative} \\
    Positive definition &
      \makecell[l]{Unguided downstream B-Rep\\fails the OCCT checker} \\
    Detector training split & None (no detector training) \\
    Development/test overlap & None \\
    Test configuration &
      \makecell[l]{Frozen before held-out\\evaluation} \\
    Evaluation target &
      \makecell[l]{Downstream invalidity-risk\\prediction} \\
    \bottomrule
  \end{tabular*}
  \caption{GTAD development and held-out test protocol. The development cohort is used only for model and threshold selection; final detector metrics are reported on the disjoint held-out test cohort after freezing all configurations.}
  \label{tab:supp-gtad-protocol}
\end{table}

\subsection{Detector and Guidance Configurations}
We render a $2\times2$ composite containing front, right, top, and isometric views. All four cameras are orthographic; the axis-aligned views use the corresponding canonical directions, and the isometric camera uses direction $(1,1,1)$. Each wireframe is centered and isotropically scaled so that its longest bounding-box side has length two, rendered as black lines on a white background with a 5\% margin, and stored as one $512\times512$ image. We query \texttt{qwen/qwen3.5-flash-02-23} through the OpenRouter Chat Completions API~\cite{openrouter2026api} using its default provider-routing policy~\cite{openrouter2026routing} and temperature 1.0. An API or JSON-parsing failure is retried once; a second failure is conservatively parsed as \texttt{fail} and therefore activates both guidance branches.

\paragraph{External VLM Data Governance.}
Each request sends only the normalized $512\times512$ multi-view wireframe rendering and the fixed detection prompt to OpenRouter; no native CAD/B-Rep file, point cloud, sample identifier, construction history, or condition metadata is transmitted. Under default provider routing, OpenRouter forwards the request to an eligible endpoint rather than to a provider pinned by us~\cite{openrouter2026routing}. OpenRouter documents that full prompt/response logging and product-use opt-ins are disabled by default, although a small number of prompts may be anonymously categorized; endpoint-specific provider retention policies may differ~\cite{openrouter2026datacollection}. Zero Data Retention and data-collection denial can be enforced at the account or request level, while ZDR may still permit transient provider-side in-memory prompt caching~\cite{openrouter2026zdr}. Because rendered views can still reveal proprietary geometry, external VLM routing carries an IP-disclosure risk for confidential industrial CAD; such deployments should enable \texttt{zdr=true} and \texttt{data\_collection=deny}, pin an approved provider, or replace the external service with a locally hosted VLM.

For tangent-point energy, we use the kernel~\cite{yu2021repulsive}
\begin{equation}
k_{\beta}^{\alpha}(p,q,T_p)
=\frac{\left\|T_p\times(p-q)\right\|^{\alpha}}
{\left\|p-q\right\|^{\beta}},
\qquad (\alpha,\beta)=(3,6),
\label{eq:supp-tpe-kernel}
\end{equation}
with the default discrete evaluation and numerical handling of the public Repulsive Curves implementation~\cite{yu2021repulsivecode}. Before evaluation, we apply the same unit-box normalization described above. Interactions within the same edge or between edges sharing a vertex are excluded, while nonlocal interactions between all other edge pairs are retained.

Concretely, let $\tilde p=2(p-c_{\mathrm{box}})/L$, where $c_{\mathrm{box}}$ is the bounding-box center and $L$ is its longest side. Each straight wireframe edge $e=(i_1,i_2)$ is treated as a segment with normalized length $\ell_e$ and unit tangent $T_e$. The edge contribution in the main-paper definition is evaluated using the endpoint quadrature
\begin{equation}
E_{tpe}(e)=
\sum_{e'\in\mathcal N(e)}
\frac{\ell_e\ell_{e'}}{4}
\sum_{i\in\partial e}\sum_{j\in\partial e'}
k_{\beta}^{\alpha}(\tilde p_i,\tilde p_j,T_e),
\label{eq:supp-discrete-tpe}
\end{equation}
where $\mathcal N(e)$ contains edges distinct from $e$ that do not share a vertex with it. The ordered edge-pair aggregation in Eq.~\eqref{eq:supp-discrete-tpe} is then averaged over edges as in the main paper.

This endpoint-based discretization is a computational risk surrogate, not a complete predicate for segment intersection, edge collapse, or legal near-contact. In particular, finite endpoint quadrature need not resolve every interior event, and a high score alone does not establish kernel invalidity. We therefore use $E_{\mathrm{geom}}$ only for risk-based routing and specify the targeted controlled cases in Tab.~\ref{tab:supp-controlled-geometry}.

For autoregressive guided regeneration, the default setting is $K=4$, sampling temperature 1.0, and top-$k=4$; only the $K/N$ sweep in Tab.~\ref{tab:supp-sampling-sweep} varies $K$ and sets top-$k=K$. The one-retry rejection-sampling variant permits one complete retry, and Best-of-$N$ generates $N=4$ complete trajectories in parallel by default with the selection rule in Eq.~\eqref{eq:supp-best-of-n}. Autoregressive experiments use equal energy weights, $\lambda_{\mathrm{geom}}=\lambda_{\mathrm{topo}}=1$.

If an energy or gradient evaluation is undefined or non-finite, GTAD conservatively activates the corresponding route; EGGTR excludes the affected autoregressive candidate or skips the affected diffusion-guidance update, falling back to the original unguided generator step when no finite guided option remains.

\subsection{Evaluation Protocol}

We follow the public DTGBrepGen environment and validity checker~\cite{li2025dtgbrepgen}, using Python 3.10.13 and \texttt{occwl} 2.1.0. A sample is counted as \emph{Valid} only when the STEP file can be read and transferred successfully, contains exactly one solid, has correctly ordered and non-self-intersecting face wires, contains no shell bad edges, and has no free or open edges.

Following BrepForge~\cite{brepforge2026}, point-cloud-conditioned reconstruction uses 3,000 ABC test geometries. The baseline and WDR use identical condition IDs and base random seeds. \emph{Valid}, CD, EMD, and F-Score are all aggregated over the complete test cohort, without filtering samples according to construction success. A failed reconstruction is assigned the worst finite CD and EMD observed in the pooled comparison and an F-Score of zero. For successfully constructed B-Reps, we uniformly sample 4,096 surface points. This failure-aware protocol measures aggregate conditional reconstruction fidelity across all attempted conditions rather than a valid-only or common-success subset. The unconditional COV, MMD, and JSD experiments retain the sampling and evaluation protocols of their corresponding baselines. Novel and Unique follow the protocol of BrepGen~\cite{xu2024brepgen}. All compared variants use the same test samples, condition $c$, and base latent or random seed $z$.

\begin{table}[t]
  \centering
  \wdrtablestyle
  \begin{tabular*}{\columnwidth}{@{}l@{\extracolsep{\fill}}cccc@{}}
    \toprule
    Dataset &
    \makecell{Invalid\\$\rightarrow$ Invalid} &
    \makecell{Invalid\\$\rightarrow$ Valid} &
    \makecell{Valid\\$\rightarrow$ Invalid} &
    \makecell{Valid\\$\rightarrow$ Valid} \\
    \midrule
    DeepCAD & 95  & 520 & 37 & 2,348 \\
    ABC     & 404 & 553 & 79 & 1,964 \\
    \bottomrule
  \end{tabular*}
  \caption{Paired validity transitions for DTGBrepGen on DeepCAD and ABC, using 3,000 matched samples per dataset. Each row uses the same condition and base seed before and after applying WDR.}
  \label{tab:supp-paired-valid-transitions}
\end{table}
As shown in Tab.~\ref{tab:supp-paired-valid-transitions}, WDR increases \emph{Valid} from $79.50\%$ to $95.60\%$ on DeepCAD and from $68.10\%$ to $83.90\%$ on ABC. The paired gains are $16.10$ percentage points (95\% bootstrap CI: $[14.67,17.53]$; exact two-sided McNemar $p=3.86\times10^{-110}$) and $15.80$ points (95\% bootstrap CI: $[14.27,17.37]$; $p=1.63\times10^{-88}$), respectively. Confidence intervals use 100,000 paired bootstrap resamples. Guided regeneration or reranking therefore converts substantially more initially invalid outputs to valid than it converts initially valid outputs to invalid on both datasets.

\section{Limitations and Future Work}
\label{sec:supp-limitations}

\paragraph{Performance Bottlenecks.}
GTAD evaluates three complementary detector branches in parallel, including multi-view rendering and a general-purpose VLM. This introduces rendering and model-inference latency and remains susceptible to false positives and false negatives under ambiguous projections or dense local structures. A false positive from any detector can trigger guided regeneration or reranking, and a VLM false positive activates both branches. EGGTR further increases inference cost through repeated candidate evaluation or guided denoising. Its recovery capability is also bounded by the underlying pre-trained generator: when a severely corrupted intermediate lies outside the generator's candidate or trajectory support, the guided process may still fail to produce a valid final B-Rep. Fig.~\ref{fig:gallery-failure}(a)--(c) summarizes these residual detector and guidance failures. Future work will explore more efficient CAD-specific screening, adaptive test-time computation, and stronger underlying generators.

\paragraph{Validity Scope.}
Our \emph{Valid} metric is an operational kernel-level check under the stated OCCT/\texttt{occwl} protocol. Passing this checker does not certify manufacturability, tolerance robustness, functional correctness, or engineering suitability, all of which require application-specific analysis beyond the scope of this work.

\paragraph{Applicability.}
The current framework assumes that a generator explicitly exposes an intermediate wireframe whose geometry and topology can be inspected and used to trigger guided regeneration before surface instantiation. It therefore applies to the studied multi-stage B-Rep generation pipelines, but does not directly support single-stage models that jointly synthesize the complete B-Rep or parametric-sequence generation paradigms in which a design is represented by modeling-command and parameter sequences without a comparable wireframe intervention point. Extending WDR to these settings will require representation-specific structural detectors and guidance interfaces that operate directly in their latent or sequence spaces. We regard both single-stage and parametric-sequence generation as important directions for future work. Our conditional evaluation covers class-label and point-cloud inputs; image-conditioned generation and other conditioning modalities are outside the present scope. Moreover, the current guidance is driven primarily by validity-oriented energies. Excessively prioritizing validity may steer a guided sample away from its input condition, as illustrated by the point-cloud-conditioned examples in Fig.~\ref{fig:gallery-failure}(d). Future work will incorporate condition-consistency metrics into the guidance objective and investigate multi-objective guidance that balances structural validity with conditional fidelity.

\section{Additional Experiments}
\label{sec:supp-additional-experiments}

\begin{figure*}[t]
  \centering
  \includegraphics[width=\linewidth]{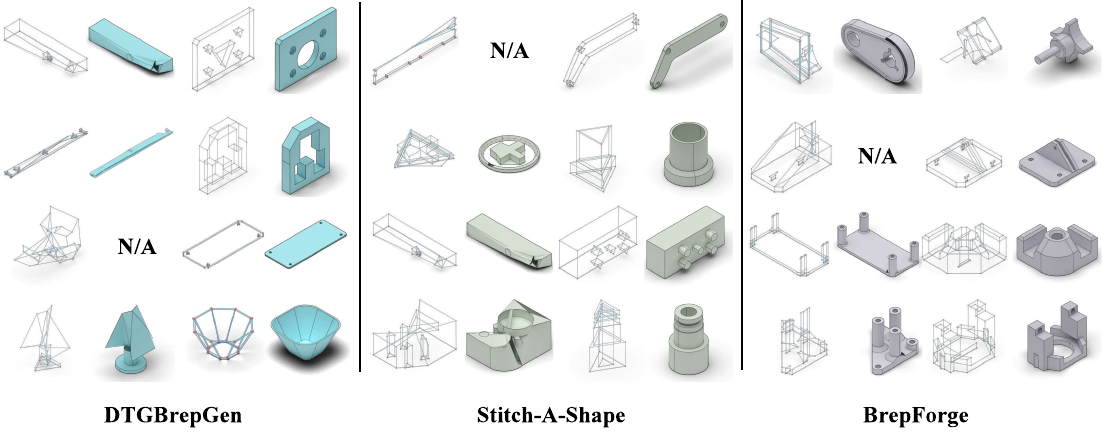}
  \caption{
  Gallery of unconditional B-Rep generation.
  We show qualitative results of \methodname on three multi-stage B-Rep generators:
  DTGBrepGen, Stitch-A-Shape, and BrepForge~\cite{li2025dtgbrepgen,li2025stitchashape,brepforge2026}.
  For each generator, each group shows the initial wireframe, the direct construction result, the WDR-guided wireframe, and its downstream B-Rep.
  "N/A" indicates construction failures from the initial anomalous wireframe, so no initial B-Rep is available for visualization.
  }
  \label{fig:gallery-uncond}
\end{figure*}

\begin{figure*}[t]
  \centering
  \includegraphics[width=0.97\linewidth]{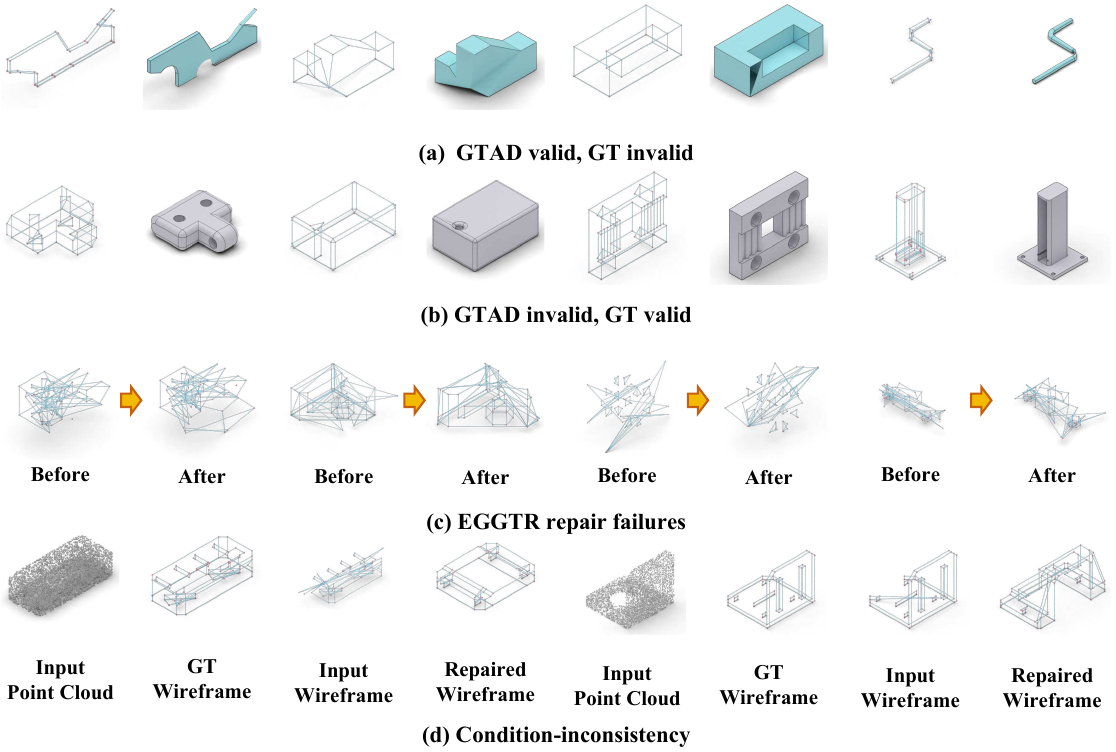}
  \caption{Representative failure cases. Here, ``GT'' denotes the downstream checker label. (a) False negatives: GTAD accepts checker-invalid wireframes. (b) False positives: GTAD flags checker-valid wireframes. (c) EGGTR guided-regeneration failures on severely corrupted wireframes. (d) Point-cloud-conditioned cases in which the guided wireframe deviates from the input condition; each group shows the input point cloud, GT wireframe, input wireframe, and guided wireframe.}
  \label{fig:gallery-failure}
\end{figure*}

Detailed class-wise \emph{Valid} scores used to compute the ten-class macro-average are reported in Tab.~\ref{tab:supp-class-valid}. Additional qualitative results for class-conditioned, point-cloud-conditioned, and unconditional generation are shown in Figs.~\ref{fig:gallery-class}--\ref{fig:gallery-uncond}. Additional screening, guided-regeneration, and condition-consistency failures are summarized in Fig.~\ref{fig:gallery-failure}.

\begin{table*}[t]
  \centering
  \wdrtablestyle
  \begin{tabular*}{\textwidth}{@{}l@{\extracolsep{\fill}}cccccccccc@{}}
    \toprule
    Method & Bathtub & Bed & Bench & Bookshelf & Cabinet & Chair & Couch & Lamp & Sofa & Table \\
    \midrule
    DTG base & 49.8 & 67.9 & 69.4 & 50.8 & 69.2 & \textbf{78.2} & 73.5 & 36.6 & \textbf{79.9} & 68.3 \\
    DTG + WDR & \textbf{67.6} & \textbf{84.0} & \textbf{80.1} & \textbf{67.4} & \textbf{75.6} & 71.5 & \textbf{76.6} & \textbf{63.9} & 78.3 & \textbf{71.3} \\
    \midrule
    SAS base & 29.7 & 70.3 & 62.5 & 53.1 & 53.1 & \textbf{60.9} & 59.4 & 60.9 & 64.1 & 71.9 \\
    SAS + WDR & \textbf{42.1} & \textbf{84.8} & \textbf{72.8} & \textbf{60.0} & \textbf{69.3} & 58.7 & \textbf{67.7} & \textbf{83.6} & \textbf{64.2} & \textbf{77.2} \\
    \bottomrule
  \end{tabular*}
  \caption{Class-wise \emph{Valid} scores (\%) on the Furniture dataset. DTG and SAS denote DTGBrepGen and Stitch-A-Shape, respectively. Bold indicates the better result within each generator group and category. The main-paper result is the unweighted macro-average over these ten categories.}
  \label{tab:supp-class-valid}
  \vspace{0.8em}
  \includegraphics[width=\linewidth]{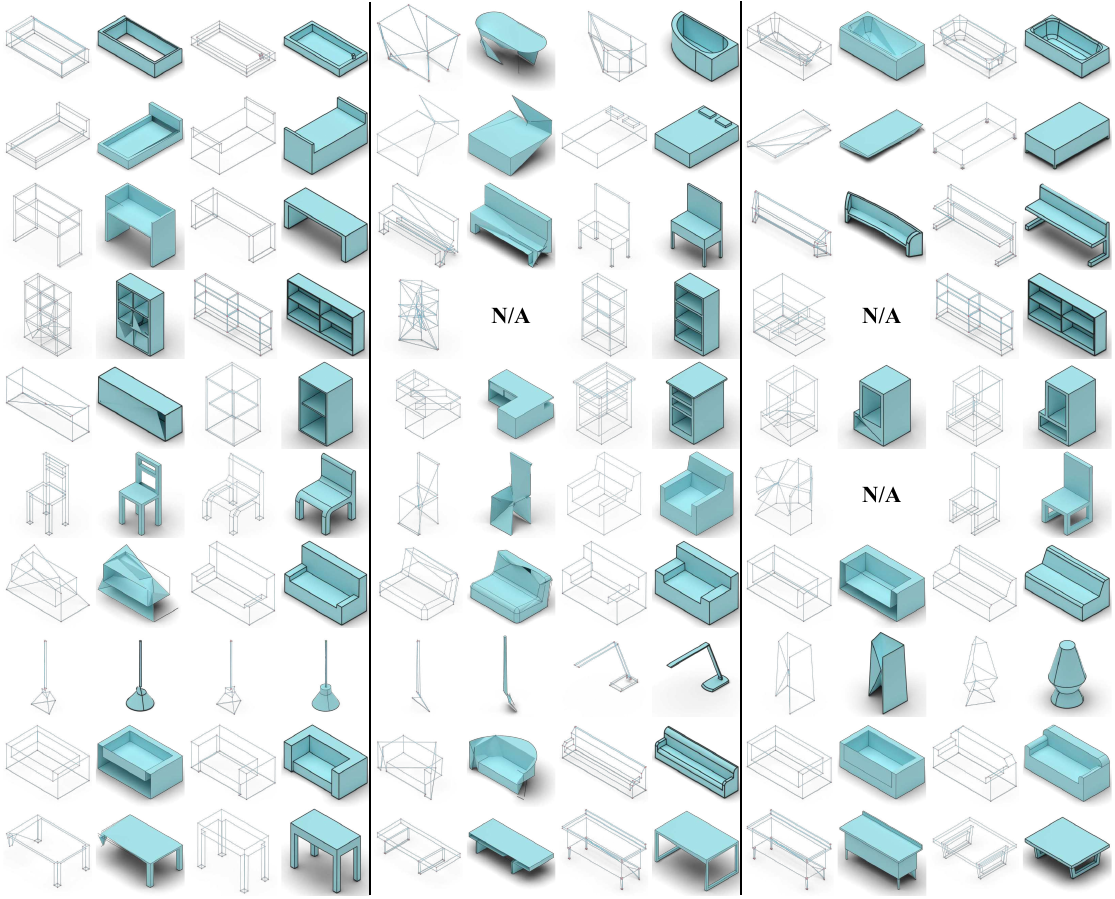}
  \captionof{figure}{
  Gallery of class-conditioned B-Rep generation.
  We show class-conditioned results with rows corresponding to Bathtub, Bed, Bench, Bookshelf, Cabinet, Chair, Couch, Lamp, Sofa, and Table from top to bottom.
  For each sample, columns from left to right show the initial wireframe associated with a failed construction, the failed B-Rep construction, the WDR-guided wireframe, and its downstream B-Rep. "N/A" indicates that the initial wireframe does not yield a B-Rep for visualization.
  }
  \label{fig:gallery-class}
\end{table*}
\begin{figure*}[t]
  \centering
  \includegraphics[width=\linewidth]{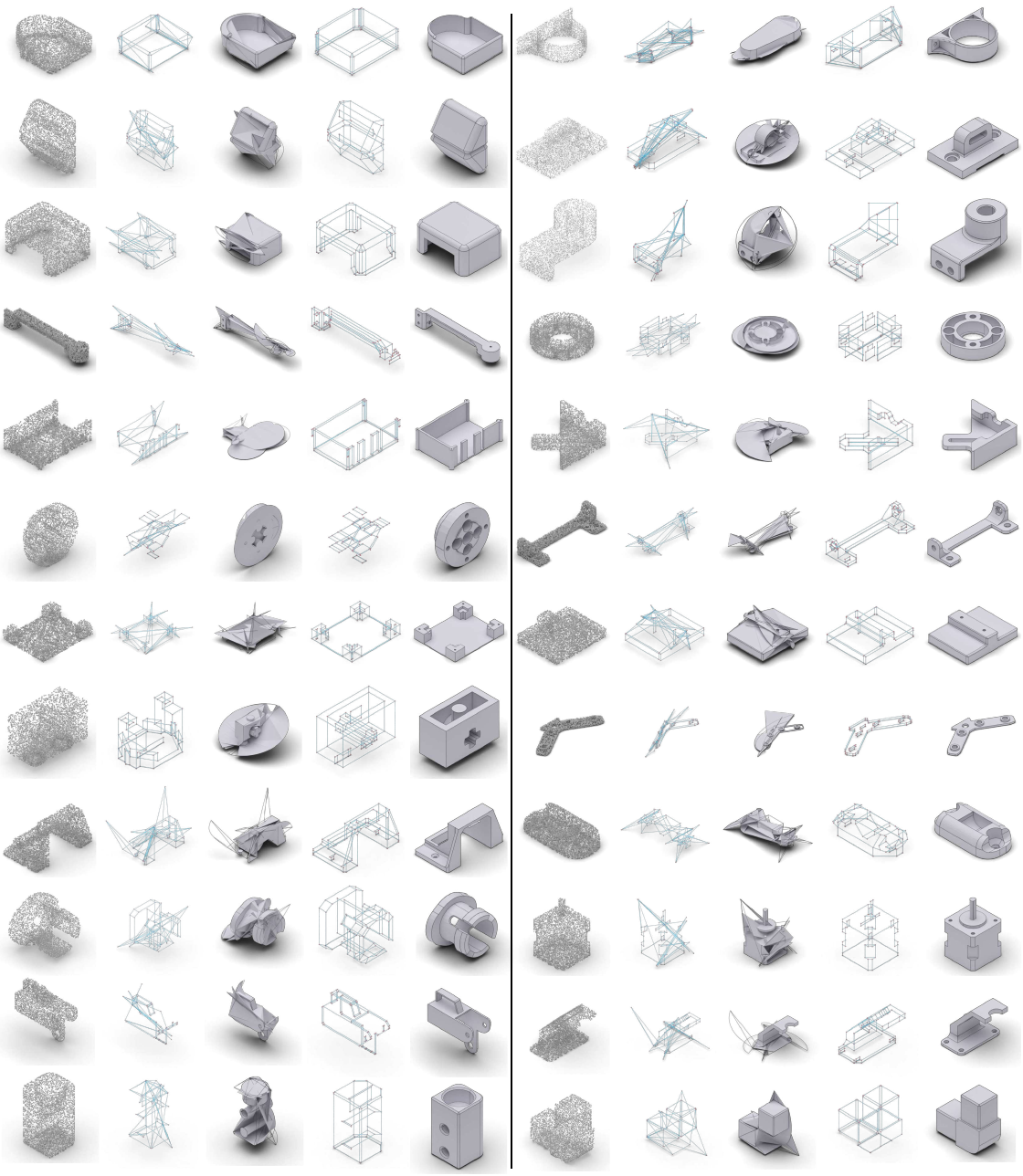}
\caption{
Gallery of point-cloud-conditioned B-Rep generation.
We show point-cloud-conditioned results on BrepForge with \methodname attached at the intermediate wireframe stage.
For each sample, columns from left to right show the input point cloud, the initial wireframe associated with a failed construction, the failed B-Rep construction, the WDR-guided wireframe, and its downstream B-Rep.
}
  \label{fig:gallery-pc}
\end{figure*}

\begin{figure*}[t]
  \begin{minipage}{\textwidth}
\section{VLM Detector Prompt}
The following prompt is used with Qwen3.5-Flash for all multi-view experiments. Each request contains the $2\times2$ composite image and the text instruction below; no additional structural metadata are supplied. The model returns only \texttt{pass} or \texttt{fail} in the displayed JSON format, without additional free-form text.
\begin{lstlisting}[style=promptstyle, caption={Prompt for multi-view wireframe anomaly screening.}, label={lst:wireframe_prompt}]
Given a 2x2 CAD wireframe image (front/right/top/isometric), judge whether the current wireframe has clear visual anomalies. Use only visible geometry. Do not infer from object category, templates, or what the object seems to be.

Output JSON only, with no markdown and no extra fields:
{"decision":"pass"} or {"decision":"fail"}

Judgment order:
1. Inspect each view locally first: line segments, endpoints, intersections, overlaps, inner/outer contours, and attachments. Then check whether the four views explain each other.
2. Output fail only for a clear, localizable, hard visual anomaly that cannot be explained by another view.
3. When a suspicious region is explained by another view, match the specific edge, endpoint, or attachment across views. Do not rely on vague explanations such as “edge-on,” “has thickness,” or “normal projection.”

Fail if any of the following is visible and not explainable by other views:
- Non-endpoint crossings, fold-backs, self-intersecting boundaries, or local edge weaving.
- Many endpoints or short edges compressed into a small region, causing unclear connection candidates, fan-outs, fold-backs, or stuck-together edges.
- Inner loops, holes, cutouts, or inner contours touch, pass through, or nearly share vertices/edges with the outer contour, making boundary ownership unclear.
- A small loop, box, cluster, or substructure looks suspended or detached in multiple views, lacks credible attachment, or has conflicting attachments.
- The same local structure disappears, breaks, interweaves, shifts, or connects differently across views, so no one-to-one correspondence can be established.
- A collapsed view cannot be explained as a clear single connection after checking the other views.
- A small box/loop/inner structure sends multiple diagonal or short candidate attachments toward the outer shell, and their endpoints cluster or cannot be uniquely paired in at least one view. Prefer fail even if the isometric view suggests a plausible 3D story.
- “Truss-like,” “frame-like,” or “edge-on” is acceptable only when endpoint pairing is clear. Still fail for converging diagonals, multiple candidate attachments, non-endpoint crossings, or highly compressed vertex heights.

Prefer pass in the following cases:
- Thin plates, rods, walls, hollow bodies, or slender frames collapse into lines in one or two orthographic views, but the remaining views or isometric view clearly explain depth, connections, and endpoint order.
- Wedges, bevels, chamfers, steps, holes, inner/outer nesting, stacked volumes, or local nesting make different views look triangular, rectangular, trapezoidal, or polyline-like, but the views remain mutually consistent.
- Simple stacking, joining, nested frames, inner/outer frame combinations, truss-like structures, or frame-like structures have many lines but no hard crossings, unreadable clusters, or attachment conflicts.
- Collinear points, parallel duplicate lines, thin rings, small closed loops, or separated-looking projections in one view are explained by thickness edges and connections in other views.
- Clean inner/outer nesting or multiple closed substructures stay consistently separated across all views, with clear spacing, no near-contact, no conflicting attachments, no dense local clusters, and no non-endpoint crossings.
- For globally simple thin plates, thin frames, or slender extrusions, collapsed front/right views with only 1–2 groups of parallel lines should pass if top/isometric views uniquely explain the contour and endpoint order. Regular repeated collinear points or double-line thickness are not dense clusters.
\end{lstlisting}
  \end{minipage}
\end{figure*}

\bibliography{references}